\documentclass[12pt]{article}
\usepackage{amsmath,amsthm,amssymb,euscript,epsf,epsfig,color}
\usepackage{array}
\usepackage{fancybox}
\usepackage[table]{xcolor}

\usepackage{hyperref}
\usepackage{todonotes}
\usepackage{subfig}
\usepackage{multirow}

\usepackage{tikz}
\usetikzlibrary{decorations.pathmorphing}
\usetikzlibrary{decorations.markings}
\usetikzlibrary{arrows.meta}
\usepgflibrary{shapes.geometric}

\newcommand{\be}{\begin{equation}}
\newcommand{\ee}{\end{equation}}
\newcommand{\bea}{\begin{eqnarray}\displaystyle}
\newcommand{\eea}{\end{eqnarray}}

\makeatletter
\@addtoreset{equation}{section}
\makeatother

\def\one{{\hbox{ 1\kern-.8mm l}}}
\def\zero{{\hbox{ 0\kern-1.5mm 0}}}

\def\tr{ \rm{tr}}

\newcommand{\ov}[1]{\overrightarrow{#1}}

  \def\cO{{\cal O}}

 \def\cZ{{\cal Z}}

\begin{document}

\rightline{QMUL-PH-22-02}
\rightline{SAGEX-22-21-E}
\vspace{1truecm}

\vspace{5pt}


{\LARGE{ 
\centerline{\bf  Permutation invariant matrix statistics} 
\centerline{\bf   and computational  language tasks } 
}}  

\vskip.3cm 

\thispagestyle{empty} \centerline{
   { \bf Manuel Accettulli Huber  ${}^{a,}$\footnote{ {\tt m.accettullihuber@qmul.ac.uk }}, }
    {\bf Adriana Correia ${}^{b,}$\footnote{ {\tt a.duartecorreia@uu.nl}}, } } 
    \centerline{ 
   { \bf  Sanjaye Ramgoolam${}^{a,c,}$\footnote{ {\tt s.ramgoolam@qmul.ac.uk}},    }
               { \bf  Mehrnoosh Sadrzadeh${}^{d,}$\footnote{ {\tt m.sadrzadeh@ucl.ac.uk }}    }
                                                       }

\vspace{.2cm}
\centerline{{\it ${}^a$ Centre for Theoretical Physics, Department of Physics and Astronomy},}
\centerline{{ \it Queen Mary University of London, Mile End Road, London E1 4NS, UK}}
    
\vspace{.2cm}
\centerline{{\it ${}^b$ Institute for Theoretical Physics and Centre for Complex Systems Studies, }}
\centerline{{\it  Utrecht University, Princetonplein 5, 3584 CC Utrecht, The Netherlands
 }}

    \vspace{.2cm}
\centerline{{\it ${}^c$ National Institute for Theoretical Physics, School of Physics and Centre for Theoretical Physics, }}
\centerline{{\it University of the Witwatersrand, Wits, 2050, South Africa } }

\vspace{.2cm} 
\centerline{{\it ${}^d$ University College London, Department of Computer Science, }}
\centerline{{\it   Gower Street, London WC1E 6BT, United Kingdom }}
\vspace{.3truecm}

\thispagestyle{empty}

\centerline{\bf ABSTRACT}

\vskip.2cm 

The Linguistic Matrix Theory (LMT) programme introduced by Kartsaklis, Ramgoolam and Sadrzadeh is an  approach to the statistics of matrices generated and used in the fields of  Computational Linguistics and Natural Language Processing. These matrices  provide semantic representations for words and phrases of natural language. 
LMT is  based on permutation invariant polynomial functions of matrices, which are regarded as  the key observables encoding the significant statistics. In this paper, we generalize the previous results on the approximate Gaussianity of matrix distributions arising from a compositional distributional approach to natural language semantics. We also introduce two geometries of observable vectors for words, defined by  exploiting the graph-theoretic basis for the permutation invariants and  the statistical characteristics of the ensemble of matrices associated with the words. We describe successful applications of this unified  framework to four tasks in semantic relation identification:  synonym versus antonym distinction, a three-way classification of synonym pairs, antonym pairs and no-relation pairs, distinction between  hypernym/hyponym pairs and co-hyponym pairs, and the distinction between a hypernym and co-hyponym within a pair.

\setcounter{page}{0}
\setcounter{tocdepth}{2}

\newpage


\tableofcontents

\section{ Introduction}

The Linguistic Matrix Theory (LMT) programme \cite{LMT,LMTQPL} proposes to use permutation invariant random matrix theories to model the statistical properties of ensembles of matrices arising from machine learning algorithms  that learn  natural language semantics in the subfields of Computational Linguistics and Natural Language Processing in Artificial Intelligence. The use of vectors to represent word meanings has a well-established history in Computational Linguistics (see for example \cite{Rieger}). This usage was  initiated in the field of \emph{distributional semantics}, the ideas behind which are  succinctly captured by J. R. Firth's famous quote ``You shall know the meaning of a word by the company it keeps'' \cite{Firth}.  Advances in  neural network machine learning in Natural Language Processing have led to  algorithms that learn meaning vectors from large corpora  of text. One such algorithm applied to mining the semantic relation of  similarity between words is word2vec \cite{MCCD1301}. Word2vec has been successfully experimented with in a variety of tasks and datasets (see e.g. \cite{Hill,Bruni}). A consistently successful measure of similarity for word2vec vectors is the cosine distance based on the Euclidean geometry in the vector space of  word meanings. In this paper, we  use the cosine in combination with length measures and experiment with a novel construction on  word vectors. 

Our  vectors are constructed from  the permutation invariant functions of matrix data learnt by extending word2vec from distributional semantics to a compositional variant of it, known as \emph{compositional distributional semantics}.   The permutation invariant functions $f(M) $ of $ D\times D $ matrices $M$  obey $f(M) = f  ( U_{ \sigma} M U_{ \sigma }^T ) $ where $U_{ \sigma }$ is the $N \times N $ matrix representing the permutations $\sigma $, and $ U_{ \sigma }^{ T} $ is the transpose. Equivalently these functions of $N^2$ variables $M_{ i j } $ ( with $i,j$ ranging over $\{ 1, 2, \cdots , N \}$), obey  $f ( M_{ ij} ) = f ( M_{ \sigma(i) \sigma (j) }) $.These permutation invariant functions play a central role in the LMT programme and are referred to as \emph{permutation invariant matrix observables} (PIMOs), which  have been studied in detail in \cite{LMT,PIGMM,GTMDS,PIG2MM}. Examples of permutation invariant polynomials of a matrix variable $M$  are 
\bea 
&& \hbox {  Linear order }  :  { \tr }  M = \sum_{ i } M_{ ii} ~~~~ \sum_{ i j } M_{ ij} \cr 
&& \hbox{ Quadratic order } : \sum_{ i , j } M_{ ij}^2 ~~~~ \sum_{ ij} M_{ ij} M_{ jj} \cdots \cr 
&& \hbox{ Cubic order } : \sum_{ i } M_{ ii}^3 , { \tr }  M^3 = \sum_{ i,j , k } M_{ ij } M_{ jk } M_{ k i } ,
\sum_{ i j } M_{ij}M_{jj}^2   , \cdots 
\eea
Some of these functions are traces, but as illustrated by the examples above, they can be more general. 
There is a general classification of matrix observables in terms of directed graphs \cite{LMT}, and the graph theoretic classification has been used to produce group-theoretic code to enumerate the graphs \cite{PIG2MM}.

Compositional distributional semantics extends representations of words from vectors to matrices and higher tensors  by modelling function words, such as adjectives or verbs, as multilinear maps. In \cite{LMT} an ensemble of adjective and verb matrices were studied. The expectation values of PIMOs  were compared with a 5-parameter permutation invariant Gaussian matrix model and we found encouraging evidence for the predictive power of the $5$-parameter model. The general $13$-parameter permutation invariant Gaussian matrix model was constructed in \cite{PIGMM} and predictions from the model were compared with the data in   \cite{GTMDS}, where  we found strong evidence for approximate Gaussianity.

The construction of matrices in \cite{LMT} was not based on  word2vec and instead a linear regression algorithm from earlier literature was used  \cite{BBZ} . In more recent work, \cite{MC2015,CSC2020} word2vec was used as a base for compositional distributional semantics. It was shown how it can be  extended to construct matrices for adjectives and verbs and how these matrices provide significant improvements over the ones learnt by linear regression. It is these latter word2vec matrices that we use in this paper. Our first result is to analyse the Gaussianity of the matrices in \cite{CSC2020} using the permutation invariant matrix model of \cite{PIGMM} and as in previous work find strong evidence for approximate Gaussianity.  This gives evidence in favour of the conjecture that approximate Gaussianity should be viewed as a universal property of ensembles of matrices that successfully represent word meanings.  The second main result is that  geometries on the PIMO vector spaces derived  using ideas from physics and statistics lead to  successful applications  in  semantic relation tasks, such as distinguishing synonyms from antonyms and hypernyms from hyponyms.

 The data analysis methods used in this paper are based on the approximate Gaussianity in the dataset. It is useful to pause here to explain the motivations from physics for the investigation of Gaussianity, which is the first focus of this paper. The construction of the matrix models in \cite{LMT,PIGMM}  use representation theoretic techniques for the combinatorics of matrix models and for matrix quantum field theories (see e.g.  \cite{CJR,KR1,BHR1,BCD0801}); these play an important role in the  Anti-de-Sitter/Conformal-Field-Theory (AdS/CFT) correspondence   \cite{AdSCFT}. The interest in Gaussian matrix distributions  in the LMT programme is  motivated by the widespread applicability of Gaussian matrix models with continuous symmetry in physics and the data sciences \cite{Wigner,EY,GMW}.  While the traditional applications are focused on matrix eigenvalue distributions, the LMT programme draws on the interpretation of matrix statistics as zero dimensional quantum field theory (QFT).  The  applicability of perturbative QFT in particle physics (see standard textbooks e.g. \cite{PS}) and in the description of cosmological perturbations (see for example \cite{SpecGauss})  naturally raises the question of whether near-Gaussian models can be applied to describe the statistics of ensembles of matrices in wider areas of data science. The discovery of the powerful role of symmetry combined with  matrix degrees of freedom in string theory \cite{AdSCFT,BFSS} in describing the quantum physics of the  simplest backgrounds of string theory, and the ideas of universality prominent in Random Matrix Theory,  give ample motivation to explore the statistics  of matrix ensembles   arising in computational linguistics and other fields.

The investigation of Gaussianity involves a comparison of theoretical expectation values calculated from the permutation invariant Gaussian matrix models and experimental expectation values. The experimental expectation values are averages obtained by summing over the different members of the ensemble of words that belong to the same Part Of Speech (POS) : in this paper,  the main focus is on different verbs in a dataset.  Denoting a permutation invariant observable  by  $\cO_{ \alpha  } ( M ) $, averages over the ensemble are of the form 
\bea 
\langle \cO_{ \alpha}  ( M ) \rangle_{expt} = { 1 \over N_{ verbs} } \sum_{ A } \cO_{ \alpha   } ( M^A )  
\eea 
$N_{verbs } $ is the number of verbs in the ensemble under consideration and $M^A$ refers to the matrix for the verb indexed by the number $A$ for $A \in \{ 1, \cdots, N_{verbs} \} $.
Since the permutation invariant polynomial functions have a basis of directed graphs, the index $\alpha$ can be regarded as running over sets of directed graphs. The first focus of the paper is to perform a Gaussianity test to understand whether the statistics of these experimentally extracted verb matrices can indeed be reproduced by the general permutation invariant 
Gaussian model. The Gaussianity  tests take as input the thirteen  linear and quadratic averages calculated from the data. These are used as input for the Gaussian matrix model and determine the thirteen parameters of the Gaussian model. The model then predicts higher order expectation values which are compared to experiment. All these steps involve averages over the set of verbs.

 Given that the  ensemble averages (over the set of verbs) of a small set (of size up to 28)  of linear, quadratic, cubic and  quartic  PIMOs allow the identification of approximate Gaussianity as an interesting feature of the randomness in the data, it is a natural  next step  to understand how values of 
the observables $ \cO_{ \alpha   } ( M^A )$ of the individual verbs indexed by $A$ are  
are situated with respect to these averages $\langle \cO_{ \alpha } ( M ) \rangle_{expt} $.    A given choice of a list of linear, quadratic, cubic and quartic PIMOs gives a   list of numbers  $ \cO_{ \alpha   } ( M^A )$ which defines a feature vector for the verbs labelled by $A$. In other words, if the ensemble averages of the PIMOs enable the identification of interesting global  characteristics of the noise in the data, it is natural to expect that the individual values of PIMOs in the ensemble also capture  some of the signals, which in this case consists of semantic information about the verbs.  This leads  to a new kind of dimensionality reduction for matrix  data allowing us to narrow the focus from a prohibitively complex data size to a manageable  set of variables which we can seek to interpret.  In the dataset of primary interest in this paper, each verb is represented by a matrix of  size $100 \times 100$. And  we have a collection of over $800 $ verbs. Studying the $10000$ matrix entries as statistical variables has a prohibitive complexity. The approach of studying the PIMOs associates, to each verb,  a vector of dimension up to $28$, which is obtained by evaluating the functions of the matrices listed in Table \ref{tab:obs}, which is introduced and discussed in more detail in section \ref{sec:gaussianity}. The motivation to focus on these specific functions rather than arbitrary functions (which would lead to very high, potentially infinite dimensionality) is based on the postulate of permutation symmetry as a useful filter for interesting and relevant semantic information in compositional distributional semantics and on the evidence for permutation invariant Gaussianity present in the low order permutation invariant polynomials. Arguments in favour of permutation symmetry are given in \cite{LMT}. Identifying the realms of applicability of actions which are quadratic (up to small cubic and quartic corrections)  in dynamical variables of physical interest (e.g. quantum fields), hence approximately Gaussian in a generalised quantum field theoretic sense,  is a widely useful  paradigm in particle physics as well as cosmology. We can thus view the experimental focus on PIMOs in this paper to be guided by the notions of symmetry and approximate Gaussianity from physics. These physics perspectives give an  approach to a very general problem of data reduction in data sciences, which is the subject of many different approaches in statistical science \cite{StatChallengeHD}.

The second focus of the present paper is then  to demonstrate that the permutation invariant polynomials also provide  valuable statistical tools for inspecting how a given matrix, corresponding to a given verb, compares to the entire ensemble of matrices. By choosing a small  set of observables (e.g. 13 or 28), we obtain observable vectors for the verbs.
We show that the geometrical and  statistical analysis of the word vectors performs successfully in a number of word comparison tasks, giving success rates significantly, i.e. by margins outside estimated errors,  above random guessing. The geometry takes the form of a choice of inner product (or metric)  on the observable vectors, and related observable deviation vectors. The first  inner product we use is physically motivated by a large N factorisation property from the physics of permutation invariant observables, a factorisation property which has recently been generalised from the case of observables with continuous symmetry \cite{BPR}. This metric is called  the \emph{diagonal} metric where  PIMOs associated to distinct directed graphs are orthogonal. We also find successful applications of a metric called the {\it Mahalanobis } metric which is a standard construction in statistics.  Our main goal in this initial step of  applications to natural language semantics is to demonstrate the viability of observable vectors and their geometries as tools for language tasks. We leave for the future the task of  fine-tuning and optimizing the metrics  and associated algorithms to obtain the best possible performance in particular tasks.

 We  apply our PIMO vectors to the task of  semantic relation identification  on the SimVerb3500 dataset \cite{SV3500}. SimVerb3500  consists of 3500 pairs of verbs labelled by different semantic relations and their associated degrees. The labels are obtained from linguistic resources and the degrees  collected from human subjects.  We use the metrics developed on the PIMO vector space to find  relationships between the average cosine values of pairs of verbs with the same semantic relation and find out that they obey an ordering consistent with  human annotations.  We use these orderings to identify the semantic relation of synonymy from antonymy and distinguish synonym pairs from antonym and no-relation pairs. The PIMO vectors and their metrics are also applied to distinguish between  hypernym/hyponym  and co-hyponym pairs, and the distinction between the hypernym and co-hyponym within a pair. We obtain a good performance in these four tasks, significantly better than metrics directly applied to word2vec vectors and to compositional distributional matrices. There exist  datasets specifically developed for distinguishing between synonymy and antonymy, hypernyms and hyponyms. In these specialised datasets, good performance has been achieved using the standard methods, see \text{e.g.} \cite{Santus,ImValde2013,AntSyn,Schwab}.  This shows that our approach is able to obtain a distinction on a dataset produced using material that is agnostic to these tasks.

The paper is organised as follows. 
Section \ref{CLback}  starts with a discussion of the challenge of identifying the synonymy/antonymy relation. 
We then describe the SimVerb3500 dataset.  Section \ref{sec:gaussianity} describes tests of Gaussianity on the datasets of matrices constructed in \cite{CSC2020}. Section \ref{sec:defs}  motivates and describes the different types of vectors of PIMOs which are associated with verbs in our discussions.  We describe two metric geometries, based on the diagonal metric and the Mahalanobis metric, and associated cosine distances for the vectors.  
  In Section \ref{sec:relations} we apply these metrics to identify synonym/antonym and hypernym/hyponym pairs of SimVerb3500. We show that observable deviation vectors, equipped with the diagonal metric or the Mahalanobis metric, consistently demonstrate small average angles between synonym pairs as compared to antonym pairs.  The distinction  of hypernyms from hyponyms is accomplished in section \ref{sec:hyper}, using the lengths of the observable deviation vectors.  Section \ref{sec:baselines} performs some baseline checks to demonstrate that simple variations of the successful constructions in sections \ref{sec:relations}, \ref{sec:precision}, \ref{sec:hyper} do not perform well in comparison. The programmes in Python developed in the course of the preparation of this paper are available at \cite{ManuelGitHub}.

\section{ Natural Language Processing models and tasks }\label{CLback}

In the 1950's,  British linguist and philosopher of language John Rupert Firth discovered a \emph{distributional} property for words and conjectured that the collocational context of words play a role  in identifying their meanings \cite{Firth}.  About the same time, American linguist and mathematician of language Zellig Harris realised the same phenomena, and formalised it in more succinct terms \cite{Harris}. Harris observed that word that have similar meanings, i.e. are synonymous,  are collocated in the context of the same words. One of his famous examples was the words  ``oculist" and  ``eye doctor",  both of which occur in the context of ``eye",  ``glasses", and ``doctor".  This school of thought is now known as \emph{distributional semantics}. Later, it was discovered that it can be used to reason about phenomena that go beyond synonymy, since words that have any semantic relationship with each other do also occur in the context of same words. Some examples are ``tea" and ``coffee", ``cat" and ``dog", and ``boy" and ``girl".  It was also observed that words that are not semantically related, e.g. ``cat" and ``coffee" do not co-occur in the same contexts. 

\begin{center}
\scalebox{0.28}{
    \includegraphics{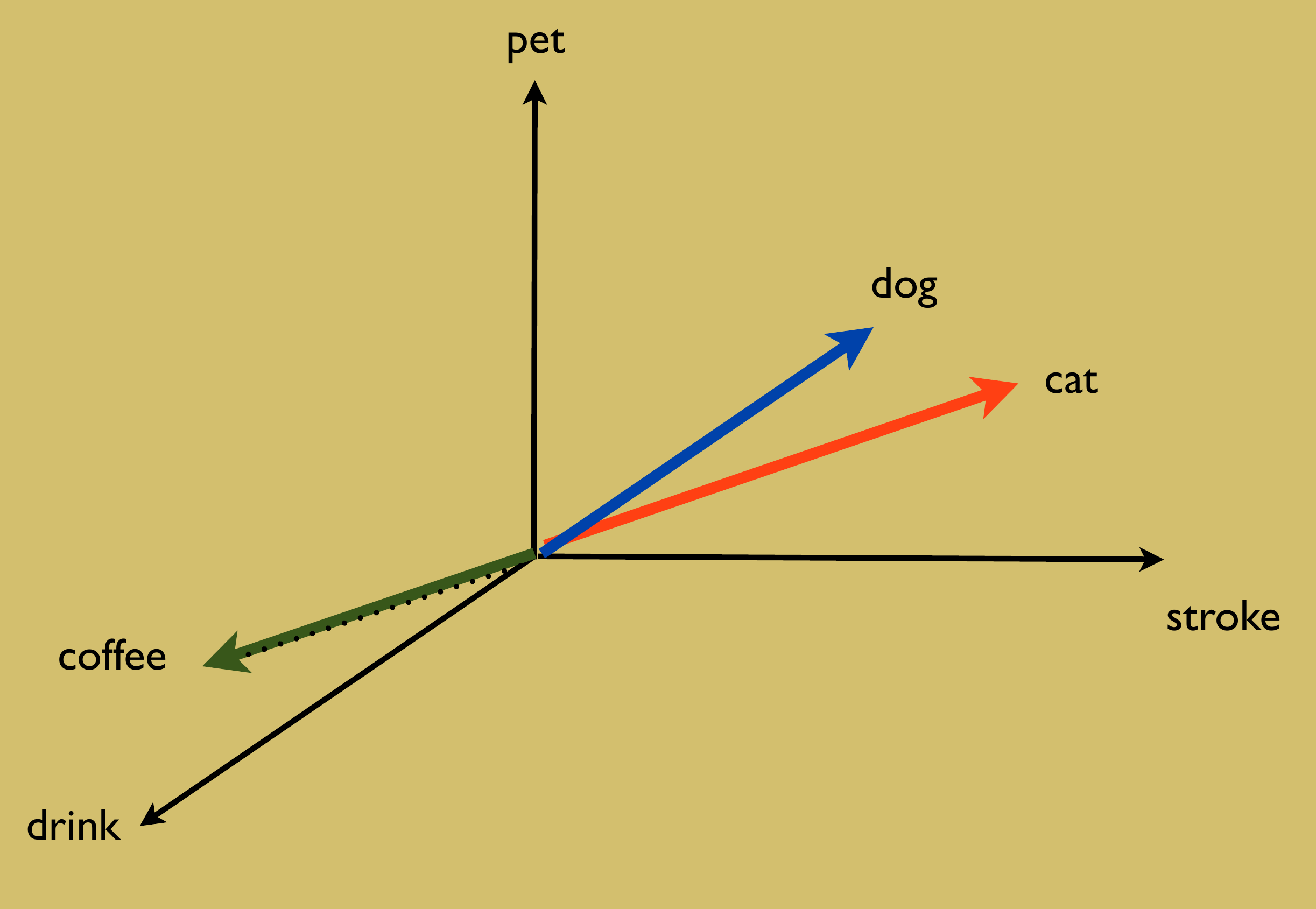}}
\captionof{figure}{  An example vector space and its word vectors }
    \label{fig:vectorspace}
\end{center}

 Distributional semantics was  implemented by the first wave of Natural Language Processing researchers which included  Rubenstein and Goodenough \cite{RubGood}. They first embedded a large corpus of natural language data into a matrix. The columns of this matrix were canonical forms of from a dictionary and were referred to as ``context". The rows denoted ``target" words and were all the words that had occurred in the corpus. Each target word was then represented by its row vector,  in a vector spaces generated by the context words as basis. As a result, words that were semantically similar got represented by vectors that were \emph{close} to each other in this space. As an example, see Figure \ref{fig:vectorspace}.

  \subsection{Synonym/Antonym Hypernym/Hyponym identification task}

Identifying similar/dissimilar pairs of words have been the stronghold of distributional semantics and many word similarity  datasets have been developed to showcase this ability. Examples of these datasets are various extracts of the  TOEFL test \cite{Resnik}, MEN \cite{Bruni}, WordSim-353 \cite{Finkelstein}, SimLex-999 \cite{Hill}, and SimVerb3500 \cite{SV3500}.   It has turned out that the cosine of the angle between  pairs of vectors  is a  good approximation of   their degree of similarity.  Fine graining  the degrees of semantic similarity  into more precise  lexical relationships of antonymy, synonymy,  hypernymy and hyponymy  or co-hyponymy has  nevertheless proved to be less straightforward. 

In this regard, methods that do not use external resources or post processing lead to unintuitive results, e.g. see \cite{Bernardi2019} where it is shown that the negation of an adjective is closer to the adjective itself rather than to its antonym. As a result,  researchers were led to develop supervised, unsupervised  and semisupervised learning methods  to obtain solutions. In the supervised approaches, we have the work of \cite{Turney} and \cite{Mohammad}, who  use thesauri categories  as labels  and run machine learning classification algorithms to learn  the relationship in unseen data.  The un- and semi-supervised methods learn the specific linguistic and non-linguistic  properties of word vector entries  and use these  to distinguish between  lexical  relations. Examples of  properties for the synonym/antonym distinction  are    salient   feature  \cite{Santus} and close or opposite features  \cite{ImValde2013,Schwab}. For the hypernym/hyponym relationship, entropy calculations  over  vector representations and their cosines are used  \cite{SantusHyper}.  Deciding what  these specific properties are has been a challenge. An example theory uses Kempson's work \cite{Kempson}, who  characterises the salient feature as a  property that holds for one word but fails for the other, e.g. the property of being tall for a giant and a dwarf. Another theory is  Cruse's \cite{Cruse}, who argues that a pair of antonym words share all of  their properties except for this one. Identifying this sole salient features  requires extensive  computations. For  opposing  relations,  instead of using a simple adjacency  hypothesis for counting co-occurrence, researchers use lexical patterns  \cite{SchulteandKoper} such as ``X as opposed to Y", ``X or Y",  and discourse markers such as ``though", ``however", ``nevertheless"    \cite{RothandSchulte}. Learning these patterns and markers requires large amounts of data  not to lead to sparsity and suffers from low coverage as the methods only works  for  data that has frequently co-occurred with these specific patterns. Furthermore, building the corresponding vector spaces requires parsing the corpora to discover the patterns, which is also costly.

The machine learning classification algorithms mentioned above are now replaced by neural network learning algorithms,   the first most widely used of which is the \emph{skipgram} algorithm\cite{MCCD1301,MK2013}; for instance \cite{Ono} uses thesaurus categories to enhance the skipgram   algorithm  and  \cite{Nguyen} integrates  lexical contrast information from WordNet into skipgram for the same purpose. More advanced   learning algorithms that  use neural networks are also been  developed and explored \cite{VuMr2018}.   These combine external linguistic constraints with neural word embeddings.  The shortcomings of all of these methods, however,  is that they rely on   manual labour  or external  resources   for labelling. Further, there is  low coherence between these   resources; WordNet-based methods have a very low overlap with  approaches that use thesauri or human annotations.  Our method, using low-dimensional vector representations and tailored similarity measures is able to consistently improve on these baseline results.

Finally, judging whether a pair of words are synonyms or antonyms is  a hard task for human annotators \cite{SV3500,Santus}. Often annotators are provided  with a sub class of words that share specific properties and nudged towards the salient feature that distinguishes them, in order to produce correct answer.  For instance, they are told that they should consider relationships such as \emph{gradable opposites}, when facing pairs of words such as (\emph{short, tall}) or \emph{complements} when facing (\emph{hot,cold}) \cite{Santus}. Another approach is to tell the annotators  to annotate the antonym pairs with a low degree of similarity, leading to unreliable inter-annotator  agreement scores that are then  dismissed from similarity evaluations \cite{SV3500}.

 \subsection{Verb similarity task}

 The SimVerb3500 dataset was developed as a  large similarity and relatedness dataset for verb pairs, aimed at being representative of the many different semantic and syntactic relationships held between verbs. Similarly to other previous word-similarity datasets, it aims to be a representative, clearly defined,  consistent dataset, that is, it encloses the full range of concepts that emerge in natural language, distinctively states the annotated relation, such as synonymy or co-hyponomy, and native english speakers should be able to identify this same relation consistently when given simple indications.

 The dataset contains 3500 verb-pairs, composed of 827 distinct verbs. The majority of the pairs were  chosen from the USF dataset \cite{USF}. Some top verb classes of the largest online verb lexicon for English, i.e. VerbNet  \cite{VN1,VN2}, were not sufficiently represented in USF. These were added to provide a representative set  of verbs of English.  The annotators were asked to rank all of the 3500 verb-pairs according to how similar the verbs within a pair were. This resulted in a similarity score for each verb-pair,  ranging from $0$ (least similar) to $10$ (most similar). A set of lexical relations were then extracted from  WordNet \cite{WN}, to  label  each verb-pairs, as  either of ``synonyms'', ``antonyms'', ``co-hyponyms'', ``hypernym/hyponym" or ``no relation''.

  In this study we consider  similarity scores and lexical relations, both of which we aim to predict.  To this effect, we slice the SimVerb3500 dataset into smaller datasets indexed by their lexical relation label. We observe in this procedure that the SimVerb3500 is not a balanced dataset in these labels, e.g. there are around 3 times more synonyms pairs (306) than antonym pairs (111). This follows naturally from the fact that the dataset was built to be representative of other kinds of relationships, namely those present in the USF datasets and VerbNet, but by using the lexical annotations we are given a window to study the distribution of these labels on a rich verb-pair dataset that was tailored to depict the many nuances of verb usage.

\subsection{Compositional distributional semantics and  verb matrices}\label{sec:VerbMatConst}  

Compositional distributional semantics (CDS) extends distributional semantics from words to phrases and sentences \cite{CSC2010,GS2015,BBZ,KSP,GDZhSB,MCG2014,MC2015,CCS2013,PPhB014}. In order to do so, it first analyses the grammatical structure of the phrase/sentence, then decides what kind of representation to build for each word therein.   The grammatical structure is analysed using a type or categorial logic, where, nouns are assigned atomic  grammatical types. In CDS, they are represented as vectors. Adjectives are  functions that modify nouns to produce adjective-noun phrases.  Verbs are functions that modify  nouns to produce sentences. In CDS, they  are represented as matrices. Some words have higher order functions as grammatical types and  in CDS, they are represented as higher order tensors, such as cubes and  hypercubes.  Examples of such words are prepositions, such as \emph{from, to},  conjunctives, such as \emph{and, or}, and wh-words, such as \emph{who, when}.  For the purpose of this paper, we focus on verbs and their matrices. This allows us to use techniques from Random Matrix Theory and in particular its permutation invariant version,  developed in our previous work  \cite{LMT}. 




The matrices of CDS are learnt from large scale data, such as corpora crawled from the web and resources such as wikipedia, Google's archives of books and news, and snapshots of film scripts and daily conversations of ordinary people. A popular choice for a corpus is  the \texttt{ukWackypedia} corpus (\url{https://wacky.sslmit.unibo.it}), which is a concatenation of the English Wikipedia with a crawl of all webpages with a  uk web domain.


The first attempt to learn the CDS matrices did not use machine learning. In their paper  \cite{GS2015},  the authors argued that the matrix of a transitive verb should be the sum of the Kronecker products of vectors of its subjects and objects across the corpus. Formally, the matrix $\overline{verb}$ of a verb $verb$ is constructed as follows
\[
\overline{verb} = \sum_i \ov{Sub}_i \otimes \ov{Obj}_i
\]
for $i$ an index that goes over all the sentences of the corpus in which the verb  is $verb$ and the subject and object of the verb are $Sub$ and $Obj$. 
%

%
%
A later  approach used machine learning and implemented a linear regression algorithm to learn the matrices \cite{BBZ,GDZhSB,PPhB014}. This resulted in the  production of  higher quality matrices, which we used in previous work \cite{LMT}.

 At about the same time, artificial neural network algorithms started to dominate the field of machine learning for Natural Language Processing \cite{MCCD1301,MK2013}. Similar to other mathematical optimisation  algorithms, artificial neural networks   try to select the best available  values of a feature or a set of features (here represented by a vector)  by maximising or minimising a real function, called an \emph{objective function}.  What is learnt by these algorithms is the weights of the edges between the nodes of the network. Seminal work presented in \cite{MCCD1301} showed how one can learn  efficient vector representations for words by using only a one layer neural network algorithm.  This architecture came to be known as \emph{skipgram}. The skipgram representations were learnt by maximising a logistic regression objective function by maximising the inner product between a set of target words and their contexts.  Later, in \cite{MK2013}, the authors showed that if at the same time as maximising the inner product between a target word and its context one also minimises the inner product between the same target words and word that are \emph{not} in its context, the vector representations acquire a much higher quality. This algorithm came to be known as  \emph{skipgram with negative sampling}.

 Given a target word $n$, a set of words in its context $c \in C$, are referred to as \emph{positive samples}.  A set of words not in its context $\overline{c} \in \overline{C}$, are referred to as \emph{negative samples}. The context of a  word consists of all the other words in a  window of $k$ words around it. Deciding what $k$ should be is subject to experimentation, often it is a number between $3$ and $10$, within the same sentence boundaries. The skipgram with negative sampling algorithm learns representations for  target and context words as  vectors  $n, c$ and $\overline{c}$.  Initially, these vectors  are randomly initialised and as skipgram with negative sampling proceeds with learning, they are simultaneously updated to maximise the  objective function of the algorithm, given below:

\begin{equation} \label{objective}
 \sum_{c\in C} \log \sigma(\textbf{n} \cdot \textbf{c}) + \sum_{\bar{\textbf{c}} \in \bar{C}} \log \sigma(- \textbf{n} \cdot \bar{\textbf{c}}), 
\end{equation}

\noindent
Here,  $\sigma$ is the logistic function,  defined by the following formula
\[
\sigma(x) = \frac{1}{1 + e^{-x}}
\]

\noindent
 This function is differentiable and has  a non-negative derivative.  Its characteristic is an $S$-shaped curve, also referred to by a \emph{sigmoid}.  Sigmoids smoothen their input into a form that starts from bounded smaller values, and gradually progress towards a maximum.

 The skipgram with negative sampling algorithm was extended from vectors to matrices for adjectives in \cite{MC2015} by  changing its objective function to the following
\begin{equation} \label{objectiveadj}
 \sum_{c\in C} \log \sigma((\overline{\textbf{adj}} \times \textbf{n}) \cdot  \textbf{c}) + \sum_{\bar{\textbf{c}} \in \bar{C}} \log \sigma(- (\overline{\textbf{adj}} \times \textbf{n}) \cdot  \bar{\textbf{c}}), 
\end{equation}

\noindent
Similar to neural word embeddings learnt by this algorithm, the adjective matrices acquired superior performance in Natural Language Processing tasks.

In \cite{CSC2020}, the authors showed that for the case of verbs $\textbf{verb}$, with a subject $\textbf{sub}$ and an object $\textbf{obj}$  we need to learn two matrices, and thus need to have two objective functions. These are as follows:

\begin{equation} \label{objectiveverbsubj}
M_O\colon \sum_{\textbf{subj}\in C} \log \sigma((\overline{\textbf{verb}} \times \textbf{obj}) \cdot  \textbf{subj}) + \sum_{\bar{\textbf{subj}} \in \bar{C}} \log \sigma(- (\overline{\textbf{verb}} \times \textbf{obj}) \cdot  \bar{\textbf{subj}}), 
\end{equation}
and one when the context is $\textbf{obj}$, as follows
\begin{equation} \label{objectiveverbsobj}
M_S\colon \sum_{\textbf{obj}\in C} \log \sigma((\overline{\textbf{verb}} \times \textbf{subj}) \cdot  \textbf{obj}) + \sum_{\bar{\textbf{obj}} \in \bar{C}} \log \sigma(- (\overline{\textbf{verb}} \times \textbf{subj}) \cdot  \bar{\textbf{obj}}), 
\end{equation}

One of the matrices is learnt  from its subject contexts and the other one from its object contexts.  We refer to these    as  \textit{Verb-object matrix}, in short  $M_O$, obtained with subjects used as context, and  \textit{Verb-subject matrix}, in short  $M_S$, where  objects are used as context. The linear combination of these two
\begin{equation} \label{eq:mixing}
  M=aM_O + (1-a)M_S  
\end{equation}
provided a single high quality matrix for the verb and   outperformed all previous representations in the NLP tasks considered in \cite{CSC2020}, as well as the ones considered in \cite{PPhB014}.  The optimal weights, i.e. the values of $a$, are set by trial  and error for each verb, through a roughly equally spaced set of $a$ values $\{ 0 , 0.2,  0.5, 0.8, 1.0 \}$ across the range previously considered in \cite{CSC2020}.

All of the methods listed above are generalisable  to words with higher order functions. Since our focus in this paper is on the verbs, we will not  go through their  corresponding learning formulae and refer the reader to the literature listed above in the text. A final remark is that in the original CDS paper \cite{CSC2010}, intransitive verbs were assigned matrices and transitive verbs cubes. Learning  the cubes from data turned out to be computationally  costly and the resulting cubes did not have a high quality. That is why in the later papers of the field, referenced above, both intransitive and transitive verbs were learnt as matrices.

\section{  Approximate Gaussianity in machine-learned matrices for verbs   }\label{sec:gaussianity}

 A Gaussian distribution for one variable is described by a probability density function 
\bea\label{formulaGauss}  
 f ( x  ; \mu , \sigma ) = e^{ - S (x ; \mu , \sigma )  }
 \eea 
where $ S ( x ; \mu, \sigma )  ={ 1 \over 2 \sigma^2 }  ( x - \mu )^2 $. $S$ is a polynomial function of degree $2$ in the random variable $x$, depending on parameters $\mu, \sigma $ which determine the distribution.  $\mu$ is the mean and $\sigma $ is the standard deviation of the distribution.  Such probability distribution functions give a good fit to the statistical distribution of many quantities of interest in statistics. For example the heights of adult women in Brazil in 1996 was well described by a Gaussian distribution with mean 155.7 centimetres and a standard deviation of 6 cm \cite{HeightOfNations}. Given a sample of heights $x^{(A)} $, with $A$ ranging over $\{ 1, 2, \cdots , N \}$ where $N $ is a large  sample size, the mean, which we denote as $ < x >_{expt} $,  is calculated from the data as 
\bea 
< x >_{expt} =  { 1 \over N  } \sum_{ A =1}^{ N } x^{(A)}  
\eea 
This can be viewed as experimental data which determines the theoretical mean $\mu$   appearing in the formula \eqref{formulaGauss}. Higher moments are calculated 
as 
\bea 
< x^k >_{expt} =  { 1 \over N  } \sum_{ A =1}^{ N }  ( x^{(A)} )^k  
\eea
When an ensemble $\{ x^{ (A) } \}$ of real world data is well-described by a Gaussian distribution, the experimental moments are very well approximated  by the theoretical moments 
\bea 
< x^k >_{theor} = \int dx ~f (  x ; \mu , \sigma ) ~x^k =  \int dx ~ e^{ - S ( x ; \mu , \sigma ) } ~ x^k 
\eea
These are expectation values of $x^k$ obtained by integrating $x^k$ with  the measure $dx ~ e^{ - S ( x ; \mu , \sigma )}$. The function $S ( x ; \mu , \sigma )$ is called an action because of the analogies between integrals  which appear in statistics and path integrals which generalise ordinary integrals and define quantum mechanics and quantum field theory.

In the paper \cite{PIGMM}, which was motivated by results in \cite{LMT}, Gaussian probability distributions for a set of random variables $X_{ij}$, organized as a matrix with $1 \le  i , j \le D$, were considered. In particular, Gaussian distributions constrained to be invariant under a permutation symmetry, which had been argued to be relevant to compositional distributional semantics in \cite{LMT}, were described in generality. Permutation invariant Gaussian distributions are defined by a probability distribution over matrix variables $X_{ ij }$ determined by an action which generalizes $ S ( x ; \mu , \sigma )$ to an action of the form $ S ( X_{ij}  ; \mu_1 , \mu_2 , \lambda_1 , \cdots , \lambda_{11} ) $. This matrix action depends 
on the random matrix variables $X_{ij}$ and is a quadratic polynomial function of these variables. The constraint of permutation symmetry allows two parameters $\mu_1 , \mu_2 $ which are coefficients of linear permutation invariant functions of $X_{ij}$, and eleven parameters $\lambda_{ 1} , \cdots , \lambda_{ 11} $ which are coefficients of quadratic permutation invariant functions of $X_{ij}$. 
The action of the general permutation invariant model is given in equation (2.72) of \cite{PIGMM}, while the partition function is 
\bea 
\cZ  = \int dM ~ e^{ - S } 
\eea
as in eqn. (2.71) of \cite{PIGMM}. The definition of the moments of permutation invariant polynomials $f(M )$, as given in (2.73) of \cite{PIGMM}, has the form 
\bea\label{exptheor} 
\langle  f(M) \rangle_{theor} = { 1 \over \cZ }  \int dM ~ e^{ - S }  f ( M )  \,. 
\eea
 The precise form of equation (2.72)  of \cite{PIGMM} is dictated by the representation theory of the symmetric group $S_D$, as explained in \cite{PIGMM}.
The techniques for computing (2.73) described in \cite{PIGMM} are based on Wick's theorem, which is widely used in quantum field theory.  The key point for what follows here is that \cite{PIGMM} explains how to  calculate, in terms of the $13$ parameters of the permutation invariant distribution, the expectation values of permutation invariant polynomial functions of arbitrary degree. For example the formulae (3.31) and (3.32) in \cite{PIGMM} give the expectation value $ \langle \sum_{ i,  j } (  X_{ ij} X_{ ij} ) \rangle $.

The paper \cite{GTMDS} applied the $13$-parameter permutation invariant  Gaussian matrix model to study the statistics of  matrix data  constructed in \cite{LMT} and  representing a collection of adjectives. The thirteen parameters of the  model were calculated using the experimental averages of  two linear and eleven quadratic permutation invariant polynomial functions of the matrix variables. With the thirteen parameters determined, the theoretical model predicts, based on the assumption that the data has permutation invariant Gaussianity, the expectation values of the higher order polynomial functions of the matrix. The ratios of these theoretical values to the experimental averages of the corresponding permutation invariant functions were calculated and found, in a majority of cases, to be close to $1$, thus giving evidence that the matrix data indeed has, in approximate form,  the postulated Gaussianity. The system of equations which are solved to give the thirteen parameters is given in section 2.1 of \cite{GTMDS}. Analytic expressions for ten cubic/quartic expectation values are given in Appendix A of \cite{GTMDS}. The derivation of the analytic expressions in section 2.1 and Appendix A of \cite{GTMDS} are done using the results from \cite{PIGMM}. The solution of the system of equations in section 2.1 of \cite{GTMDS}is done with Python, using as input from data the linear and quadratic expectation values appearing there. The calculation of the higher order theoretical expectation values and comparison with the experimental expectation values is also done in Python. The code for these computations is available on GitHub \cite{GTMDS-GitHub} and has also been incorporated in the code for this paper \cite{ManuelGitHub}.

The experimental averages  over an ensemble of matrices for adjectives (constructed using linear regression in \cite{LMT}), of low order permutation invariant polynomial functions, analysed with the tools of permutation invariant random matrix theory, reveal the interesting structure of approximate Gaussianity. In this section we will extend the investigation of Gaussianity to the case of matrices for verbs constructed,  as reviewed in Section \ref{sec:VerbMatConst}, using neural net methods. We will find that approximate permutation invariant Gaussianity continues to hold. The existence of interesting approximate permutation invariant Gaussianity in ensembles of words, motivates the consideration of the values of low order permutation invariant polynomial functions for the individual members of the ensembles, as an interesting characteristic of the individual words with potential applications to lexical semantics. This second theme of the paper is investigated in subsequent sections. For both of these parts of the paper, it is useful to introduce the notion of observable vectors associated with words. This is a specific instance of the general idea of a feature vector in data science \cite{FeatureVector}.  The bases of the vector spaces we consider come from lists of observables which  we now specify.

 Let $ \cO_{ \alpha } ( M )$ be a permutation invariant polynomial function of a matrix $M$, which we refer to as 
a permutation invariant matrix observable (PIMO). The label  $\alpha$ is an index for the observables. 
For example we can choose
\begin{equation}
  \def\arraystetch{1.5}
  \begin{array}{c}
    \cO_{ 1 } ( M ) = \displaystyle\sum_{ i } M_{ ii} \\
    \cO_{ 2} ( M ) = \displaystyle\sum_{ i , j } M_{ i  j }  \\
  \cO_{ 3 } ( M ) = \displaystyle\sum_{ i } M_{ ii}^3  \\
  \vdots
  \end{array}
\end{equation}
In the  $13$-parameter Gaussian theory \cite{PIGMM}  $M$ is a matrix random variable.  The theoretical expectation values are defined as in \eqref{exptheor} as 
\bea\label{expobservth}  
\langle  \cO_{ \alpha} ( M )  \rangle_{theor} = { 1 \over \cZ }  \int dM ~ e^{ - S } 
\cO_{ \alpha } ( M )  
\eea
In our dataset, we have an ensemble of matrices $M^{ (A)  } $, where $ A$ runs over $ \{ 1 , 2 , \cdots , N_{ verbs } \} $ and $N_{verbs} $  is the number of verbs. As mentioned earlier, for every verb we will consider five different matrix representations corresponding to different values of the parameter $a$ in \eqref{eq:mixing}, thus we will have five different ensembles of matrices each with $N_{ verbs} $ matrices.

\begin{figure}[!htbp]
  		\centering
  		\includegraphics[width=.2\linewidth]{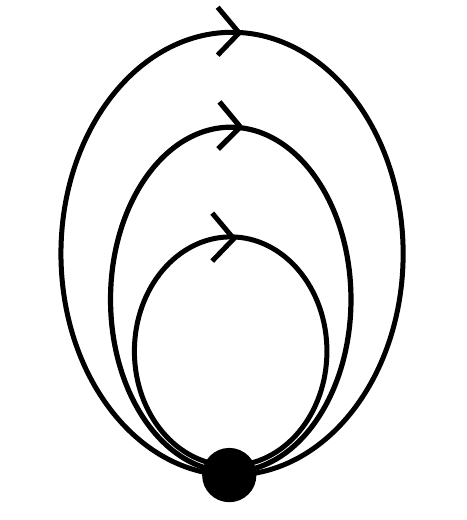}
 		 \caption{$\sum_{i} M_{ii}^{3}$}
  	\label{fig:Graph1}
\end{figure}

\begin{figure} 
	  	\centering
 		 \includegraphics[width=.4\linewidth]{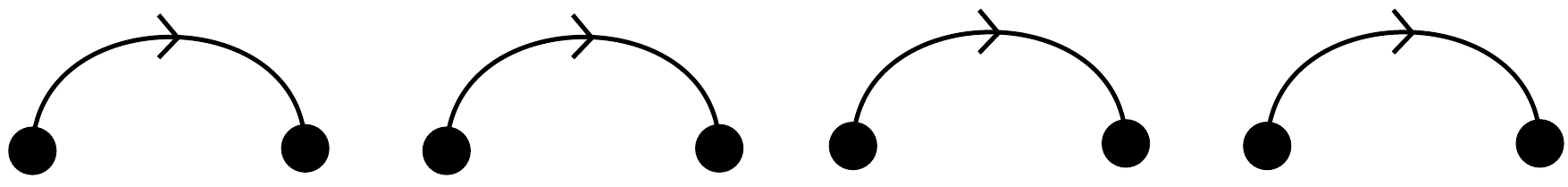}
	 	 \caption{$\sum_{i_{1}, \dots ,i_{8}} M_{i_{1}i_{2}}M_{i_{3}i_{4}}M_{i_{5}i_{6}}M_{i_{7}i_{8}}$ }
	 \label{fig:Graph10}
	\end{figure}

The complete set of observables used in this paper is provided in Table \ref{tab:obs}. The first $13$ are the linear and quadratic observables. They form a basis for the most general permutation invariant matrix polynomial of degree up to $2$, i.e. any such polynomial is a linear combination with constant coefficients (which can be  taken to be general real numbers) of these $13$ PIMOs. The next $10$ are a selection of cubic and quartic PIMOs. There is a graphical representation of the observables, where the summed indices correspond to nodes and $M_{ij}$ corresponds to a directed edge going from node $i$ to node $j$. For cubic and quartic observables, we can have between one and  $8$ nodes.  Observable 14 in Table \ref{tab:obs}   ( $\sum_{ i } M_{ ii }^3 $ )  has one summed index and a corresponding directed graph with one node (Figure \ref{fig:Graph1}), while Observable 23    ( $\sum_{ i,j, k, l , m, n, o , p } M_{ ij} M_{ kl} M_{ mn} M_{ op} ) $   has eight summed indices and a corresponding graph with eight nodes (Figure \ref{fig:Graph10}).   These $10$ are a diverse subset in the sense of including graphs with $1,2,3,4,5,6,7,8$ nodes. The theoretical computations of the expectation values of these observables were done by hand in \cite{PIGMM,GTMDS}. Computer code for general cubic, quartic and higher order observables was given in \cite{PIG2MM} (the code can be downloaded from the arXiv repository for the paper). To verify the robustness of the Gaussianity result in \cite{GTMDS} we computed the extra  $5$ observables in Table \ref{tab:obs}.

\begin{table}[h]
  $$\begin{array}{c|c}
    \# & \text{Observable} \\
    \hline
     &   \mbox{Linear and Quadratic}\\
     \hline 
    1 &  M_{ii} \\
    2 &  M_{ij} \\
    3 & M_{ij}M_{ij} \\
    4 & M_{ij} M_{ji} \\
    5 & M_{ii}M_{ij} \\
    6 & M_{ii}M_{ji} \\
    7 & M_{ij}M_{ik} \\
    8 & M_{ij}M_{kj} \\
    9 & M_{ij}M_{jk} \\
    10 & M_{ij}M_{kl} \\
  \end{array}
  \hspace{0.5cm}
  \begin{array}{c|c}
    \# & \text{observable} \\
    \hline
    11 & M^2_{ii} \\
    12 & M_{ii} M_{jj} \\
    13 & M_{ii}M_{jk} \\
    \hline
    & \mbox{Cubic and Quartic }\\
    \hline
    14 & M_{ii}^3 \\
    15 & M^3_{ij} \\
    16 & M_{ij}M_{jk}M_{ki} \\
    17 & M_{ij}M_{jj}M_{jk} \\
    18 & M_{ij}M_{kk}M_{ll} \\
    19 & M_{ij}M_{jk}M_{ll} \\
    \multicolumn{2}{c}{\textcolor{white}{space}}
  \end{array}
  \hspace{0.5cm}
  \begin{array}{c|c}
    \# & \text{observable} \\
    \hline
    20 & M_{ij}M_{kl}M_{mm} \\
    21 & M_{ij}M_{kl}M_{mn} \\
    22 & M_{ij}M_{kl}M_{mn}M_{oo} \\
    23 & M_{ij}M_{kl}M_{mn}M_{op} \\
   \hline
    24 & M^4_{ii} \\
    25 & M^4_{ij} \\
    26 & M_{ij}M_{jk}M_{pq}M_{qr} \\
    27 & M_{ij}M_{jk}M_{kl} \\
    28 & M_{jk}M_{kl}M_{lm} \\
    \multicolumn{2}{c}{\textcolor{white}{space}}
  \end{array}$$
 \caption{ The complete list of the observables used in this paper. Summations over every index are understood. The first 13 observables are the linear and quadratic observables. Observables 14-28 are a diverse selection of cubic and quartic observables used to verify Gaussianity. Details are  given in the text.}
  \label{tab:obs}
\end{table}

It is useful to define observable-value vectors for words as 
\bea\label{eq:ObservValVec} 
v^{A}_{ \alpha }  =  \cO_{ \alpha }  ( M^{ A} ) 
\eea
$A$ is a label for the words, while $\alpha $ is a label running over a set of permutation invariant matrix polynomials. $\cO_{ \alpha }  ( M^{ A} ) $ is the evaluation of the polynomial $\cO_{ \alpha } ( M ) $ for the matrix $M^A$ associated with the $A$'th word. 
It is also useful to   define  observable-deviation vectors as follows  
\bea \label{eq:obsdevvec}
( \delta v)^A_{ \alpha }  = \cO_{\alpha }  ( M^{A}  ) - \langle \cO_{ \alpha }  ( M ) \rangle_{  expt} \> ,
\eea
where 
\bea\label{expobservexp}  
\langle \cO_{ \alpha }  ( M ) \rangle_{  expt} = { 1 \over N_{verbs} } \sum_{ A } \cO_{ \alpha }  ( M^A  ) \> .
\eea
The  squared standard-deviation for the observable $ \cO_{ \alpha } ( M  ) $ can be expressed in terms of the observable deviation vector:

\bea 
(\sigma_{ \alpha} ( M  ) )^2 
  = { 1 \over N_{ verbs} } \sum_{ A }  ( \cO_{ \alpha } ( M^A ) - \langle \cO_{ \alpha }  ( M ) \rangle_{ expt}  )^2 \rangle 
= { 1 \over N_{ verbs} }  \sum_{ A }   ( ( \delta v)^A_{ \alpha }  )^2  .
\eea
This gives a measure of the spread in values of the observable labelled $ \alpha $, across the ensemble of verbs. The standard deviation itself is obtained by the square root :  
\bea\label{expobservdev}  
\sigma_{ \alpha} ( M  ) 
= \sqrt  { {1 \over N_{ verbs} } \sum_{ A }   ( ( \delta v)^A_{ \alpha }  )^2  }  \, . 
\eea

\begin{center}
  \begin{table}[h]
    $$
    \rowcolors{2}{gray!45!white}{gray!20!white}
  \begin{array}{l|c|c|c|c|c}
  \text{Observables} &  1 & 0.8 & 0.5 & 0.2 & 0 \\
    \hline
 M_{ii}^3 &   0.206 & 0.200 & 0.057 & 0.050 & 0.065 \\
  M_{ij}M_{jk}M_{ki} &    0.053 & 0.018 & 0.078 & 0.097 & 0.102 \\
 M_{ij}M_{jk}M_{ki} &    0.024 & 0.064 & 0.088 & 0.071 & 0.018 \\
 M_{ij}M_{jj}M_{jk}  &    0.067 & 0.032 & 0.118 & 0.138 & 0.140 \\
M_{ij}M_{kk}M_{ll} &     0.055 & 0.032 & 0.026 & 0.070 & 0.073 \\
M_{ij}M_{jk}M_{ll} &     0.020 & 0.061 & 0.013 & 0.039 & 0.039 \\
M_{ij}M_{kl}M_{mm} &     0.079 & 0.084 & 0.060 & 0.007 & 0.015 \\
M_{ij}M_{kl}M_{mn}  &     0.086 & 0.024 & 0.049 & 0.082 & 0.089 \\
M_{ij}M_{kl}M_{mn}M_{oo} &     0.068 & 0.021 & 0.016 & 0.026 & 0.043 \\
M_{ij}M_{kl}M_{mn}M_{op}  &     0.093 & 0.059 & 0.048 & 0.078 & 0.085 \\
 M^4_{ii}  &     0.169 & 0.166 & 0.159 & 0.135 & 0.131 \\
 M^4_{ij}  &    0.173 & 0.181 & 0.170 & 0.156 & 0.155 \\
 M_{ij}M_{jk}M_{pq}M_{qr}  &    0.132 & 0.166 & 0.162 & 0.057 & 0.048 \\
M_{ij}M_{jk}M_{kl} &     0.012 & 0.048 & 0.062 & 0.061 & 0.052 \\
M_{jk}M_{kl}M_{lm} &     0.068 & 0.022 & 0.050 & 0.055 & 0.041 \\  
  \end{array}
    $$
    \caption{Normalised difference values for the 15 cubic and quartic observables considered in Table \ref{tab:obs}. Each column corresponds to different values of the parameter $a$ in \eqref{eq:mixing}. }
    \label{tab:gaussianity1}
  \end{table}
  
\end{center}

We now consider how well the  13-parameter permutation invariant Gaussian matrix model \cite{PIGMM}  describes the permutation invariant statistics of the collection of matrices for the SimVerb3500 dataset constructed in  \cite{CSC2020}. 
The strategy is similar to the one used in  \cite{GTMDS}, where Gaussianity was tested for an ensemble of matrices constructed in \cite{LMT} by using a linear regression method. The ensemble averages of the two linear PIMOs and of the eleven quadratic PIMOs are matched with those from the $13$-parameter Gaussian model. The parameters of the model are fixed to reproduce these expectation values. The model is then used to compute the expectation values of cubic and quartic PIMOs. For the first ten of the cubic/quartic PIMOs in Table \ref{tab:obs} the theoretical computations are available from \cite{PIGMM,GTMDS}. For the additional five, we used the computer code available  from \cite{PIG2MM}. This approach used in fitting the Gaussian model to the data is the method of moments in statistics
\cite{Wikimoments}. It is also close in spirit to the application of path integrals in particle physics. The action of the standard model of particle physics has terms in low orders (up to quartic)  in the field variables much as the Gaussian model has terms of up  to quadratic order. The matrix Gaussian integral can be viewed as a finite analogue of the  quantum field theory (QFT) path integral. The higher order moments here are analogous to  QFT  Green's functions  with three or four QFT operators. 

While the comparisons in \cite{GTMDS} were done by comparing ratios of theoretical to the experimental expectation values, this measure can be problematic when the experimental expectation values are very small. In this case, small differences can lead to ratios significantly far from one. A more reliable measure of Gaussianity is given by taking the magnitude of the difference between the theoretically predicted and the experimentally calculated, divided by the standard deviation of the experimental values of the observable. Quoting experimental values along with their standard deviations as a measure of the experimental uncertainty is of course standard practice in testing theories in experimental science as well as industrial applications (see for example \cite{Wikistd}). 
  We will thus tabulate the quantity 
\begin{equation}\label{eq:normdif}
  \bigg|\frac{\langle \cO_{\alpha}  ( M ) \rangle_{ expt } - \langle \cO_{\alpha}  ( M ) \rangle_{ theor}}{ 
\sigma_{ \alpha} ( M )     }\bigg| \>.
\end{equation}
as a measure of departure from Gaussianity,  where
$ \langle \cO_{\alpha}  ( M ) \rangle_{ expt },   \sigma_{ \alpha} ( M )     $ and $\langle \cO_{\alpha}  ( M ) \rangle_{ theor} $  are as defined in 
equations \eqref{expobservexp}, \eqref{expobservdev}, \eqref{expobservth}.   The closer this normalized  difference is to zero, the better the matrix model captures the behaviour of the data and the higher the Gaussianity. 
In Table \ref{tab:gaussianity1} we give the values of \eqref{eq:normdif} for the fifteen observables from number 14 to 28 as given in Table \ref{tab:obs}. The largest difference is $20.6 \% $ of the experimental standard deviation. For a large majority ($56$ out the $75$ entries, i.e.  $75\% $ of the entries, in the table) the difference is less than $10 \%$ of the standard deviation.  This shows that the Gaussian model gives very good predictions for the cubic and quartic moments. 

For the sake of comparison with previous results \cite{LMT,GTMDS}, where a Gaussianity comparison was performed through ratios of experimental vs theoretical means, we also compute \eqref{eq:normdif} for the dataset considered in that paper. This corresponds to matrices describing a set of adjectives and verbs in a vector space of dimension $D$.  In Table \ref{tab:gaussianity2} we provide results for $D=\{ 300,700,1300,2000\}$\footnote{Our tables include five additional observables with respect to the ones discussed in \cite{GTMDS}, see the Section 3 there for the relevant tables.}. We again find that a significant number  of observables ($57$ out of $120$ or approximately $47\%$ show  differences between theoretical and experimental values which are less that $ 10\% $ of the experimental standard deviation. The largest departures from Gaussianity are within about one standard deviation. Again, the overall conclusion is that the $13$-parameter model gives very good predictions for the cubic and quartic observables. Considering the fine structure of the non-Gaussianities we may conclude that the dataset in \cite{LMT}, which was constructed by traditional linear regression, shows overall a larger non-Gaussianity than the dataset constructed using neural networks in \cite{CSC2020}. The main conclusion to be drawn is that  approximate permutation invariant Gaussianity in ensembles of matrices constructed in compositional distributional semantics has a robust range of validity including different methods of constructing the matrices. This can be viewed a new  manifestation of the notion of universality which is familiar from random matrix theories \cite{GMW}.


\begin{center}
  \begin{table}
    $$
    \rowcolors{2}{gray!45!white}{gray!20!white}
  \begin{array}{c|c|c|c|c}
    & \text{jj 300} & \text{jj 700} & \text{jj 1300} & \text{jj 2000}
   \\
   \hline
  M_{ii}^3 &  0.238 & 0.229 & 0.273 & 0.307 \\
  M_{ij}M_{jk}M_{ki} &   0.192 & 0.653 & 0.884 & 1.039 \\
  M_{ij}M_{jk}M_{ki} &   0.020 & 0.765 & 1.160 & 1.375 \\
  M_{ij}M_{jj}M_{jk}  &  0.054 & 0.242 & 0.307 & 0.366 \\
 M_{ij}M_{kk}M_{ll} &  0.011 & 0.024 & 0.026 & 0.022 \\
 M_{ij}M_{jk}M_{ll} &  0.052 & 0.002 & 0.041 & 0.040 \\
 M_{ij}M_{kl}M_{mm} &   0.014 & 0.016 & 0.017 & 0.011 \\
 M_{ij}M_{kl}M_{mn}  &   0.024 & 0.021 & 0.022 & 0.015 \\
 M_{ij}M_{kl}M_{mn}M_{oo} &   0.037 & 0.037 & 0.038 & 0.025 \\
 M_{ij}M_{kl}M_{mn}M_{op}  &   0.047 & 0.042 & 0.043 & 0.031 \\
  M^4_{ii}  &   0.306 & 0.268 & 0.322 & 0.349 \\
 M^4_{ij}  &   0.304 & 0.207 & 0.326 & 0.382 \\
  M_{ij}M_{jk}M_{pq}M_{qr}  &  0.145 & 0.080 & 0.118 & 0.121 \\
 M_{ij}M_{jk}M_{kl} &   0.031 & 0.215 & 0.246 & 0.296 \\
 M_{jk}M_{kl}M_{lm} &  0.076 & 0.185 & 0.279 & 0.348 \\  
  \end{array}
  \hspace{0.5cm}
  \rowcolors{2}{gray!45!white}{gray!20!white}
  \begin{array}{c|c|c|c}
    \text{vb 300} & \text{vb 700} & \text{vb 1300} & \text{vb 2000}
   \\
   \hline
   0.210 & 0.239 & 0.282 & 0.308 \\
   0.011 & 0.567 & 0.884 & 1.053 \\
   0.164 & 0.335 & 1.095 & 1.271 \\
   0.084 & 0.198 & 0.265 & 0.295 \\
   0.003 & 0.021 & 0.012 & 0.010 \\
   0.007 & 0.021 & 0.039 & 0.040 \\
   0.010 & 0.013 & 0.007 & 0.005 \\
   0.032 & 0.022 & 0.023 & 0.030 \\
   0.035 & 0.032 & 0.023 & 0.023 \\
   0.052 & 0.038 & 0.040 & 0.050 \\
   0.281 & 0.278 & 0.326 & 0.343 \\
   0.307 & 0.205 & 0.311 & 0.361 \\
   0.109 & 0.068 & 0.117 & 0.135 \\
   0.043 & 0.166 & 0.207 & 0.266 \\
   0.119 & 0.143 & 0.236 & 0.300 \\  
  \end{array}
    $$
    \caption{Values of \eqref{eq:normdif} for the matrix dataset used in \cite{GTMDS} with the four values $D=\{ 300,700,1300,2000\}$, on the left the results for adjectives and on the right for verbs.  }
    \label{tab:gaussianity2}
  \end{table}
\end{center}

\section{Geometry of observable vectors }\label{sec:defs}

  Our investigation of Gaussianity in section \ref{CLback} found strong evidence for approximate Gaussianity, by considering the theoretical and experimental expectation values of low order (linear up to quartic) PIMOs. This result can be viewed as revealing interesting structure in the randomness present in the ensemble of matrices representing verbs as constructed in \cite{CSC2020}. It is natural to expect that the same observables contain information about the individual matrices rather than the whole ensemble. This motivates us to look at vectors of observables as low-dimensional characteristics of the individual matrices. We recall that the matrices are of size $100 \times 100$, while a small number of observables ($13$ linear/quadratic and $15$ cubic/quartic) suffice to capture  the approximate Gaussianity.   Finding the realms of applicability of simple actions determined by symmetry is a widespread tool for understanding physics. Considering all the matrix entries as random variables involves looking at $10000$ variables.  Considering the low-dimensional vectors of observables can be viewed as a physically motivated technique for data reduction, based on the notions of permutation symmetry and Gaussianity.

Building on our discussion of Gaussianity,   we consider two types of observable vectors: we call them observable-value vectors $v_{ \alpha }( M^A ) $ from equation 
\eqref{eq:ObservValVec} and observable-deviation vectors $(\delta v)_{ \alpha}^A$ from \eqref{eq:obsdevvec}. In this section, we will consider associated inner product geometries on these vector spaces.

 The focus now turns, beyond Gaussianity tests, to the performance of these PIMOs in lexical semantics tasks.  There are in total $52$ cubic observables and $256$ observables, as explained using group theoretic arguments in \cite{LMT}. The measurements of Gaussianity as well as the experimentation with tasks can clearly be extended to include more of these observables. This paper should be viewed as an initial proof of concept for ideas, where we are applying constructions from permutation invariant random matrix theories defined in \cite{LMT,PIGMM} and ideas from QFT and particle physics to understand  the structure of matrix datasets in computational linguistics. These ideas can be explored in more detail with additional observables and more fine-tuned algorithms based on the framework here, some of which we will comment on in section \ref{sec:sumout}.

\subsection{Observable-value vectors }\label{sec:ObsVals}  

To build an observable-value vector for a given verb labelled by $A$ we define 
\bea \label{eq:obsvalvec}
v^A_{ \alpha  } = \cO_{ \alpha } ( M^A ) \> ,
\eea 
in other words, each entry of the vector is a permutation invariant object.

A natural inner product  for observable-value vectors is given by
\begin{equation}\label{eq:obsvalprod}
g_{ val}  ( v^A , v^B ) = \sum_{ \alpha  } { v^A_{\alpha }  v^B_{ \alpha } \over \langle  \cO_{\alpha }   ( M )^2  \rangle_{expt} } \> ,
\end{equation}
where 
\begin{equation} 
\langle  ( \cO_{\alpha }   ( M ) )^2\rangle_{expt} = { 1 \over N_{verbs} } \sum_{ A }  ( \cO_{\alpha }  ( M^A ))^2  \> .
\end{equation}
 Under this inner product, the different observables, corresponding to different directed graphs, are orthogonal and the square of the standard deviation is used to set the scale along the direction specified by each graph. Part of the motivation for experimenting with this inner product is the property of large $N$ (here large $D$) factorisation of observables in matrix theories. This property, familiar for observables in matrix models with continuous symmetries, was extended to permutation invariant observables in \cite{BPR}.

The number of elements defining the range of  $\alpha $ depends on a choice of a set of observables.  Let us call this  set $S$. Particular choices of $S$ considered in this paper are the  $13$, $23$ and $28$ observables following the ordering of Table~\ref{tab:obs}, as well as further sets of $10$ and $15$ elements taken to be rows $14$ to $23$ and $14$ to $28$ respectively. If we want to make this choice explicit  we will write 
\bea 
g_{ val }  ( v^A , v^B ; S ) \> .
\eea
The cosine distance between vectors is then defined as usual as
\begin{equation} \label{eq:cosdis}
\cos ( v^A , v^B ) =  { g_{ val}  ( v^A , v^B ) \over \sqrt { g_{val}  ( v^A , v^A ) g_{ val}  ( v^B , v^B ) } } \> .
\end{equation}

\subsection{Observable-deviation vectors  and diagonal metric}\label{obsdevdiag}

While the experiments testing  Gaussianity in section \ref{sec:gaussianity}  focus on ensemble averages ( sums running over the index $A$ which runs over the set of all the verbs), it will be very fruitful in the following to consider how  the observable value $\cO_{ \alpha } ( M^A ) $  of a particular verb $A$ compares to the ensemble average of that observable across the set of verbs.  We thus define  observable-deviation vectors as follows  
\bea \label{eq:obsdevvec}
v^A_{ \alpha }  = \cO_{\alpha }  ( M^A ) - \langle \cO_{ \alpha }  ( M ) \rangle_{  expt} \> ,
\eea
where 
\bea 
\langle \cO_{ \alpha }  ( M ) \rangle_{  expt} = { 1 \over N_{verbs} } \sum_{ A } \cO_{ \alpha }  ( M^A  ) \> .
\eea
The  squared standard-deviation 
\bea 
  (\sigma_{\alpha} ( M ) )^2  =   { 1 \over N_{ verbs} } \sum_{ A }  ( \cO_{ \alpha } ( M^A ) - \langle \cO_{ \alpha }  ( M ) \rangle_{ expt}  )^2 \> .
\eea
gives a measure of the spread in values of the observable labelled $ \alpha $, across the ensemble of verbs. An inner product (or metric), which treats the different directed graphs as orthogonal in line with the factorisation property \cite{BPR}, and uses the spread as a unit of distance along the direction defined by each observable, is  
\bea\label{ObsDevSig} 
g_{ dev}  ( v^A , v^B ) = \sum_{ \alpha  } { v^A_\alpha v^B_\alpha\over   (\sigma_{\alpha} ( M ) )^2   } \>,
\eea
This metric defines what we will call the diagonal  geometry for the observable-deviation vector space. The cosine distance is still defined as \eqref{eq:cosdis}. 


\subsection{Observable-deviation vectors from Gaussian matrix models}\label{obdevGauss} 

The objects introduced so far, namely the observable-value vectors \eqref{eq:obsvalvec} and the observable-deviation vectors \eqref{eq:obsdevvec}, are defined basing entirely on the empirical data. On the other hand, given the interest in Gaussianity, we may also  define another variation on the definition of the observable deviation vectors 
which uses input from the theoretical model. 
This input is formed by the  theoretically computed means which can be obtained separately for each observable. More specifically, the means of the cubic and quartic observables in a Gaussian model are related to the means of the linear and quadratic ones\footnote{For further details on this relation see \cite{GTMDS}}. This allows us to define a new observable-deviation vector as
\bea \label{eq:newdev}
w^A_{\alpha}  = \cO_{ \alpha } ( M^A ) - \langle \cO_{\alpha}  ( M ) \rangle_{ theor} \> ,
\eea
where $ \langle \cO_{\alpha }  ( M ) \rangle_{ theor} $ is the value predicted by the Gaussian model. Since the experimental linear and quadratic values are used as input for the theoretical model, for the 13 observables these are actually identical to the experimental averages
\bea 
\langle \cO_{\alpha} ( M ) \rangle_{ theor}  = \langle \cO_{\alpha} ( M ) \rangle_{ expt }  \>.
\eea
For cubic and quartic observables the theoretical values are in general different, and we expect these to be close to experiment when Gaussianity is high. In this construction, it is natural to define the scale for each observable using the average of the deviations \eqref{eq:newdev} as
\bea 
(\sigma_{ \alpha  }(M) )^2 = { 1 \over N_{ verbs} } \sum_{ A } ( w^A_{\alpha}  )^2 \> ,
\eea
and the inner product analogous to \eqref{ObsDevSig} is then
\bea \label{eq:diagonalinner}
g_{ dev}  ( w^A , w^B ) = \sum_{ \alpha   } { w^A_{\alpha }  w^B_{ \alpha } \over ( \sigma_{\alpha }  (  M ) )^2  } \>.
\eea

\subsection{Cosine distance with Mahalanobis metric }

So far the inner product of observable and observable deviation vectors was performed using an inner product which is diagonal in the basis of directed graphs.  Another approach is to define an inner product  using the inverse covariance matrix of the data. This is used in the Mahalanobis distance \cite{WikiMaha,Mahafirst}, which is designed to take into account correlations between the different observables, and is a statistical tool in anomaly detection. This is defined as 
\begin{equation}
  d_M(V^A,V^B) = \sqrt{(V^A - V^B)^T K (V^A - V^B)} \> ,
\end{equation}
where $V^A$, $V^B$ are some vectors and $K$ is the inverse covariance matrix.  Along these lines, we define the following inner product of two observable deviation vectors $v^A$ and $v^B$:
\begin{equation}\label{eq:maha}
  g_{ dev}^{\rm Maha}  ( v^A , v^B ) = \sum_{\alpha,\beta } v^A_{\alpha }  K_{\alpha \beta } v^B_{ \beta }  \> .
\end{equation}
Notice that if the observables are uncorrelated then $K_{\alpha \alpha } = 1/ (\sigma_{ \alpha}^2 ) $ and $K_{ \alpha \beta } =0$ for $\alpha \ne \beta $, thus restricting the summation to $\alpha = \beta $ in \eqref{eq:maha} which precisely reproduces the previously used inner product \eqref{ObsDevSig}. The new definition for the cosine distance becomes then
\begin{equation}
Cos( v^A , v^B )^{\rm Maha} =  { g_{ val}^{\rm Maha}  ( v^A , v^B ) \over \sqrt { g_{val}^{\rm Maha}  ( v^A , v^A ) g_{ val}^{\rm Maha}  ( v^B , v^B ) } } \> .
\end{equation}

In  the following sections we will, on occasion, present results obtained from the Mahalanobis metric for comparison  with those of the diagonal metric. We do not present detailed comparisons systematically since, upon performing the necessary computations of sections \ref{sec:relations} \ref{sec:precision}, we  find very close agreement with the simpler diagonal measure. This is a priori not obvious, especially comparing the cosine distance distributions we present in section \ref{sec:distributions}, however we always found it to be the case. There will be one case in Section  \ref{sec:hyper} where the success rate in a lexical relation task is visibly better with the Mahalanobis measure.

\section{Distinguishing lexical semantic relations using observable deviation vectors  } \label{sec:relations}

We will present  evidence in this section that the diagonal or Mahalanobis geometries of the observable-deviation vectors, defined in section \ref{sec:defs}, are good detectors of lexical semantic relations.  We will present the bulk of the discussion in terms of the definition of observable deviation vectors given in (\ref{eq:obsdevvec}), but as discussed in Section 
\ref{tasksGOBDEV} the results obtained from the definition (\ref{eq:newdev}) are similar.
Specifically, we find that the average cosines of  pairs of observable-deviation vectors obey the pattern: 
\begin{equation}\label{ANS}  
\hbox { ANTONYMS $<$  NONE $ < $ SYNONYMS } 
\end{equation}
This property is robust against a number of  changes  of the specific set $S$ of observables one considers, as we describe shortly. Following traditional applications of the cosine function of pairs of vectors in distributional semantics, larger cosines  (smaller angles) are associated with pairs that are more alike. Thus, we are finding that the observable-deviation vectors equipped with diagonal  or Mahalanobis geometry produce cosines aligned with an expectation that synonyms have higher cosines than antonyms and that no-relation pairs (labelled none in \eqref{ANS}) are intermediate. This intuition is justified by the similarity ratings  reported in SimVerb3500, where we have

\begin{center}
\begin{tabular}{lc}
Word Relation & SimVerb3500 Rating\\
\hline
Synonyms& 6.79\\
None & 3.43\\
Antonyms & 0.98
\end{tabular}
\end{center}

Results presented in this section make use of the matrix dataset of \cite{CSC2020} and the word pairs found in SimVerb3500. For the list of observables making up the vectors in each case we refer the reader to Table \ref{tab:obs}. We first present the  means, alongside the standard errors of these means,  for the cosine distances of both observable-value and observable-deviation vectors in each category, namely ``antonym'', ``synonym'' and ``none'', and for each of the five matrix representations. These values are presented only for vectors corresponding to some significant sets of observables, but it is worth mentioning that further tests have been carried out on a number of different subsets of observables, including random subsets of the 28 observables of Table \ref{tab:obs}, as well as specific subsets characterised by the number of nodes of the observables, in the sense of \cite{LMT}; results are always in line with those presented below. The pattern which emerges is that using the diagonal scalar product \eqref{ObsDevSig} or \eqref{eq:maha}   and the observable-deviation vectors the pattern \eqref{ANS} is always respected. On the other hand, we find  that \eqref{ANS} does not hold for the raw observable vectors (as defined in section \ref{sec:ObsVals}), and neither does it if we consider a plain scalar product instead of the cosine distance for the deviation-vectors, as discussed in \ref{sec:plain}. We do a number of baseline checks in section \ref{sec:baselines}.  These confirm the expectation that straightforward constructions in distributional semantics are not able to distinguish synonym pairs from antonym pairs consistently.

\subsection{The means of cosine distance of observable deviation vectors}

In this section we give explicit results for means  of the cosine distances computed for observable-deviation vectors. In particular we consider the 13 observables case \eqref{eq:means13} (the completely independent observables in the Gaussian model), the 10 \eqref{eq:means10} and 15 observables case \eqref{eq:means15} (these are some of the cubic and quartic observables of the full Gaussian model), the 23 observable case \eqref{eq:means23} and the complete set of 28 observables \eqref{eq:means28}.  Alongside each mean we quote the ``standard error of the mean'' 
\cite{Wiki-standard-error}. This is the standard deviation divided by the square root of the number of samples being averaged : i.e for the three columns in the tables below, reading from left to right,  the square root of  the number of antonym pairs (111), no-relation pairs (2093) and the number of synonym pairs (306)\footnote{The remaining pairs in SimVerb-3500 are given by 190 CoHyponyms and 800 Hyper/Hyponym pairs.}. While the standard deviation  gives the variation within the sample being averaged, the standard error is a measure of the uncertainty in the means.
\vskip.5cm 
\bea\label{eq:means13}
&& \hbox{Mean cosine distance, with errors,}\cr 
&&  \hbox{ using the $13$ linear and quadratic observables: } \cr 
 &&  \rowcolors{2}{gray!45!white}{gray!20!white}
  \begin{array}{c|c|c|c}
    a & \rm{ANTONYM} & \rm{NONE} & \rm{SYNONYM} \\
    \hline
    1 & 0.117 \pm 0.046 & 0.16 \pm 0.012 & 0.291 \pm 0.029 \\
0.8 & 0.121 \pm 0.048 & 0.166 \pm 0.012 & 0.306 \pm 0.028 \\
0.5 & 0.135 \pm 0.049 & 0.18 \pm 0.012 & 0.322 \pm 0.028 \\
0.2 & 0.142 \pm 0.049 & 0.183 \pm 0.012 & 0.321 \pm 0.028 \\
0 & 0.15 \pm 0.047 & 0.18 \pm 0.012 & 0.314 \pm 0.028 \\
  \end{array} 
\eea

\bea\label{eq:means10}
&& \hbox{Mean cosine distance, with errors,}\cr 
&&  \hbox{ using the $13$ linear and quadratic observables: } \cr 
&&   \rowcolors{2}{gray!45!white}{gray!20!white}
  \begin{array}{c|c|c|c}
    a & \rm{ANTONYM} & \rm{NONE} & \rm{SYNONYM} \\
    \hline
    1 & 0.122 \pm 0.046 & 0.267 \pm 0.012 & 0.336 \pm 0.031 \\
0.8 & 0.127 \pm 0.045 & 0.265 \pm 0.012 & 0.34 \pm 0.03 \\
0.5 & 0.187 \pm 0.045 & 0.333 \pm 0.012 & 0.383 \pm 0.029 \\
0.2 & 0.231 \pm 0.053 & 0.414 \pm 0.012 & 0.461 \pm 0.031 \\
0 & 0.296 \pm 0.053 & 0.443 \pm 0.012 & 0.478 \pm 0.031 \\
  \end{array}
\eea

\bea \label{eq:means15}
&& \hbox{Mean cosine distance, with errors,}\cr 
&&  \hbox{ using  $10$ cubic/quartic observables (as specified in the text)  } \cr 
 &&  \rowcolors{2}{gray!45!white}{gray!20!white}
  \begin{array}{c|c|c|c}
    a & \rm{ANTONYM} & \rm{NONE} & \rm{SYNONYM} \\
    \hline
    1 & 0.121 \pm 0.042 & 0.282 \pm 0.012 & 0.337 \pm 0.03 \\
0.8 & 0.144 \pm 0.043 & 0.287 \pm 0.012 & 0.359 \pm 0.029 \\
0.5 & 0.234 \pm 0.044 & 0.363 \pm 0.012 & 0.423 \pm 0.028 \\
0.2 & 0.256 \pm 0.047 & 0.422 \pm 0.012 & 0.454 \pm 0.03 \\
0 & 0.3 \pm 0.049 & 0.445 \pm 0.012 & 0.464 \pm 0.03 \\
  \end{array}
\eea

\vskip.6cm 

\bea\label{eq:means23}
&& \hbox{Mean cosine distance, with errors,}\cr 
&&  \hbox{ using the $15$ cubic/quartic observables (as specified in the text)  } \cr 
&&  \rowcolors{2}{gray!45!white}{gray!20!white}
  \begin{array}{c|c|c|c}
    a & \rm{ANTONYM} & \rm{NONE} & \rm{SYNONYM} \\
    \hline
    1 & 0.093 \pm 0.042 & 0.168 \pm 0.011 & 0.284 \pm 0.027 \\
0.8 & 0.092 \pm 0.042 & 0.172 \pm 0.011 & 0.295 \pm 0.027 \\
0.5 & 0.124 \pm 0.045 & 0.187 \pm 0.011 & 0.31 \pm 0.027 \\
0.2 & 0.132 \pm 0.045 & 0.196 \pm 0.011 & 0.319 \pm 0.027 \\
0 & 0.148 \pm 0.043 & 0.196 \pm 0.011 & 0.314 \pm 0.027 \\
  \end{array}
\eea

\bea\label{eq:means28}
&& \hbox{Mean cosine distance, with errors,}\cr 
&&  \hbox{ using the $28$ observables (as specified in text) } \cr 
 && \rowcolors{2}{gray!45!white}{gray!20!white}
  \begin{array}{c|c|c|c}
    a & \rm{ANTONYM} & \rm{NONE} & \rm{SYNONYM} \\
    \hline
    1 & 0.087 \pm 0.04 & 0.177 \pm 0.011 & 0.281 \pm 0.027 \\
0.8 & 0.093 \pm 0.041 & 0.181 \pm 0.011 & 0.298 \pm 0.027 \\
0.5 & 0.129 \pm 0.044 & 0.199 \pm 0.011 & 0.316 \pm 0.027 \\
0.2 & 0.135 \pm 0.044 & 0.205 \pm 0.011 & 0.32 \pm 0.027 \\
0 & 0.151 \pm 0.042 & 0.204 \pm 0.011 & 0.315 \pm 0.027 \\
  \end{array}
\eea

As can be seen form the data the ordering \eqref{ANS} is always preserved, independently of the five choices of the set $S$  of observables defining the vectors we consider, and independently of the choice of the value 
 of the parameter $a$ (equation (\ref{eq:mixing}))  from the construction of the matrices . It is interesting to see that considering the Mahalanobis measure \eqref{eq:maha}, the values of the means change drastically but the ordering is still preserved.  The tables show that the separation of the means between synonyms, no-relation pairs and antonyms are typically at least  twice as large as  the standard errors. As an example in the table \ref{eq:means28}, for $a=1$ the separation of the means in the first two entries of the first row is $0.177-0.087=0.09$ which is $2.25$  times  $0.04$ (standard error for the antonyms) and nine  times $0.01$ (the standard error for the no-relation pairs).   The separation of $2.25$ standard errors implies a confidence interval of 97.5 \% following a standard Gaussianity assumption in statistical inference \cite{confint}.     It is important to distinguish the standard deviation of the samples (a measure of the variation in the different word pairs) from the standard error in the means themselves. The significance of the separation of the means in units of the standard error of the means is compatible with the   robustness of the ordering  of means
we find by varying the set of observables or the $a$-parameter. This   motivates a classification method for verb pairs based on the cosines.

\begin{equation}\label{eq:meansMaha}
  \rowcolors{2}{gray!45!white}{gray!20!white}
  \begin{array}{c|c|c|c}
    a & \rm{ANTONYM} & \rm{NONE} & \rm{SYNONYM} \\
    \hline
    1 & -0.005 \pm 0.023 & 0.075 \pm 0.007 & 0.141 \pm 0.018 \\
0.8 & 0.02 \pm 0.022 & 0.083 \pm 0.007 & 0.159 \pm 0.018 \\
0.5 & 0.045 \pm 0.028 & 0.097 \pm 0.007 & 0.161 \pm 0.02 \\
0.2 & 0.044 \pm 0.028 & 0.09 \pm 0.008 & 0.148 \pm 0.021 \\
0 & 0.049 \pm 0.029 & 0.082 \pm 0.008 & 0.136 \pm 0.021 \\
  \end{array}
\end{equation}

\begin{figure}
  \centering
  \includegraphics[scale=0.4]{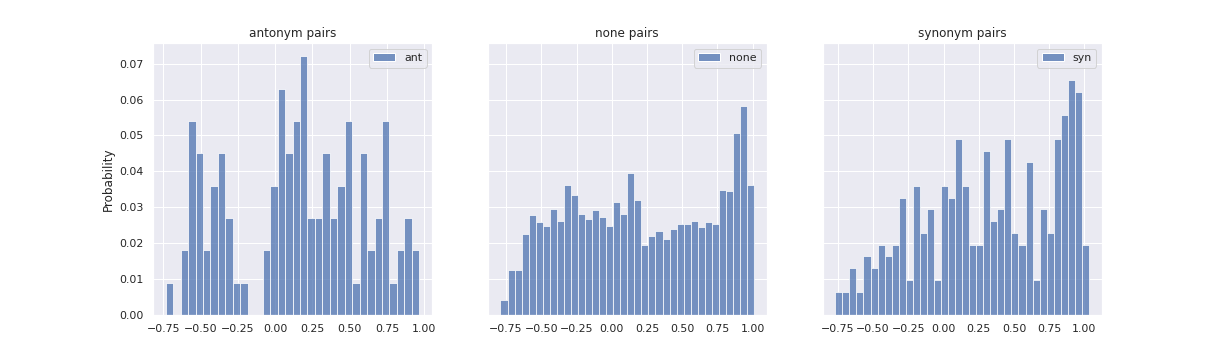}
  \caption{Distribution of the cosine distances constructed from the  \emph{diagonal inner product} of  pairs
  of observable deviation vectors  in SimVerb3500 divided by lexical relation, Antonym pairs to the left, Synonym pairs to the right and pairs with no relation in the middle. The distributions are normalised to one. Here we have considered the full 28 observable case.}
  \label{fig:distributions1}
\end{figure}

\begin{figure}
  \centering
\includegraphics[scale=0.4]{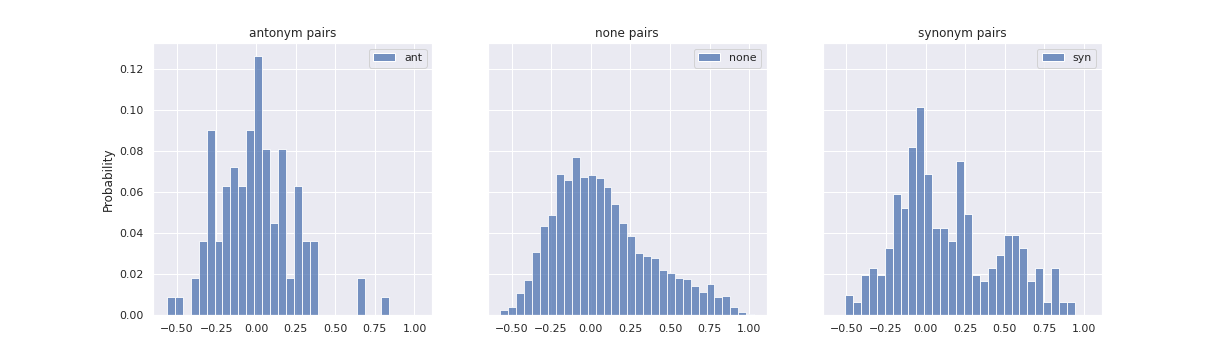}
  \caption{Distributions of the  cosine distances of the \emph{Mahalanobis inner product} of pairs of  observable deviation vectors in the 28 observable case. The three plots present the cosine distance for the SimVerb3500 word pair divided by lexical relation, Antonyms to the left, Synonyms to the right and no relation (None) in the centre.}
  \label{fig:distributions2}
\end{figure}

The non-triviality of this robustness is further highlighted by the very different shapes of the cosine distance distributions for the diagonal product and the Mahalanobis product, these are plotted in Figure \ref{fig:distributions1} and Figure \ref{fig:distributions2} respectively. These differing shapes lead to differences in success rates in the classification tasks which we discuss in Section \ref{sec:precision}.

\subsection{About observable-value vectors and other scalar products}\label{sec:plain}

It is important to remark that the combination of observable-deviation vectors with the definition of cosine distance \eqref{eq:cosdis} is rather unique, among the various variations of this choice, in producing the consistent intuitive ordering in \eqref{ANS}.   Considering for example the values of the means computed from observable-value vectors with the inner product \eqref{eq:obsvalprod} the ordering  does not hold. This can be seen from \eqref{eq:obsvalmean}, which corresponds to 28 observables, but we checked it to be true for all the combinations of observables given in Table \ref{tab:obs}.
\begin{equation}\label{eq:obsvalmean}
  \rowcolors{2}{gray!45!white}{gray!20!white}
  \begin{array}{c|c|c|c}
    a & \rm{ANTONYM} & \rm{NONE} & \rm{SYNONYM} \\
    \hline
    1 & 0.641 \pm 0.035 & 0.592 \pm 0.008 & 0.594 \pm 0.02 \\
0.8 & 0.556 \pm 0.034 & 0.606 \pm 0.007 & 0.603 \pm 0.019 \\
0.5 & 0.615 \pm 0.033 & 0.589 \pm 0.007 & 0.631 \pm 0.018 \\
0.2 & 0.576 \pm 0.036 & 0.59 \pm 0.007 & 0.595 \pm 0.02 \\
0 & 0.66 \pm 0.036 & 0.586 \pm 0.008 & 0.615 \pm 0.02 \\
  \end{array}
\end{equation}

In a similar way we can see that even favouring the observable-deviation vectors over the observable-value vectors, but opting for an inner product different from  \eqref{ObsDevSig} or  \eqref{eq:maha},
the ordering breaks down again. Here  we considered the following: 
\bea \label{ObsDevFlat} 
g_{ dev}  ( v^A , v^B ) = \sum_{ \alpha   } { v^A_{\alpha }  v^B_{ \alpha }  } \> .
\eea
which is diagonal in the graph basis for the observables, but does not take account of the standard deviations as \eqref{ObsDevSig} does. 
Computing again the means, we find for 13 observables
\begin{equation}
  \rowcolors{2}{gray!45!white}{gray!20!white}
  \begin{array}{c|c|c|c}
    a & \rm{ANTONYM} & \rm{NONE} & \rm{SYNONYM} \\
    \hline
    1 &  0.145 \pm 0.039 &  0.194 \pm   0.016 & 0.359 \pm  0.065 \\
    0 & 0.302 \pm 0.042 & 0.320 \pm 0.016  & 0.466 \pm  0.064 \\
  \end{array}
\end{equation}
For 23 observables:
\begin{equation}
  \rowcolors{2}{gray!45!white}{gray!20!white}
  \begin{array}{c|c|c|c}
    a & \rm{ANTONYM} & \rm{NONE} & \rm{SYNONYM} \\
    \hline
    1 &  0.437 \pm 0.050 &  0.578 \pm   0.017 & 0.557 \pm  0.077 \\
    0 & 0.675 \pm 0.042 & 0.644 \pm 0.017  & 0.693 \pm  0.068 \\
  \end{array}
\end{equation}
For 28 observable:
\begin{equation}
  \rowcolors{2}{gray!45!white}{gray!20!white}
  \begin{array}{c|c|c|c}
    a & \rm{ANTONYM} & \rm{NONE} & \rm{SYNONYM} \\
    \hline
    1 &  0.436 \pm 0.050 &  0.578 \pm   0.017 & 0.557 \pm  0.076 \\
    0 & 0.675 \pm 0.042 & 0.644 \pm 0.017  & 0.693 \pm  0.068 \\
  \end{array}
\end{equation}
We see that the nice behaviour of \eqref{ANS} is in general absent, the only exception being the 13 observable case. This is to be compared with the previous results for the cosine distance which does not present any breakdown of the ordering of the means.

\subsection{Word relations from Gaussian observable-deviation vectors}\label{tasksGOBDEV} 

Finally we consider the case in which the observable-deviation vectors are defined through \eqref{eq:newdev}, in other words when these vectors are built making use of the added theoretical knowledge coming from the Gaussian matrix model. In the following we present the cosine distances computed using \eqref{eq:diagonalinner} for 23, 28, 10 and 15 observables respectively in \eqref{eq:th23}, \eqref{eq:th28}, \eqref{eq:th10} and \eqref{eq:th15}.
\begin{equation}\label{eq:th23}
  \rowcolors{2}{gray!45!white}{gray!20!white}
  \begin{array}{c|c|c|c}
    a & \rm{ANTONYM} & \rm{NONE} & \rm{SYNONYM} \\
    \hline
 1 & 0.089\pm 0.042 & 0.163\pm 0.011 & 0.280\pm 0.027 \\
 0.8 & 0.090\pm 0.042 & 0.170\pm 0.011 & 0.293\pm 0.027 \\
 0.5 & 0.123\pm 0.045 & 0.186\pm 0.011 & 0.308\pm 0.027 \\
 0.2 & 0.123\pm 0.045 & 0.190\pm 0.011 & 0.312\pm 0.027 \\
 0 & 0.138\pm 0.043 & 0.188\pm 0.011 & 0.306\pm 0.027 \\
  \end{array}
\end{equation}
\begin{equation}\label{eq:th28}
  \rowcolors{2}{gray!45!white}{gray!20!white}
  \begin{array}{c|c|c|c}
    a & \rm{ANTONYM} & \rm{NONE} & \rm{SYNONYM} \\
    \hline
    1 & 0.084\pm 0.040 & 0.167\pm 0.011 & 0.275\pm 0.027 \\
    0.8 & 0.092\pm 0.041 & 0.174\pm 0.011 & 0.294\pm 0.026 \\
    0.5 & 0.124\pm 0.044 & 0.189\pm 0.011 & 0.308\pm 0.026 \\
    0.2 & 0.123\pm 0.044 & 0.192\pm 0.011 & 0.308\pm 0.027 \\
    0 & 0.138\pm 0.042 & 0.189\pm 0.011 & 0.302\pm 0.027 \\
  \end{array}
\end{equation}

\begin{equation}\label{eq:th10}
  \rowcolors{2}{gray!45!white}{gray!20!white}
  \begin{array}{c|c|c|c}
    a & \rm{ANTONYM} & \rm{NONE} & \rm{SYNONYM} \\
    \hline
    1 & 0.053\pm 0.043 & 0.180\pm 0.011 & 0.261\pm 0.029 \\
    0.8 & 0.067\pm 0.042 & 0.169\pm 0.011 & 0.262\pm 0.028 \\
    0.5 & 0.065\pm 0.044 & 0.174\pm 0.011 & 0.225\pm 0.028 \\
    0.2 & 0.035\pm 0.045 & 0.205\pm 0.011 & 0.239\pm 0.029 \\
    0 & 0.111\pm 0.050 & 0.225\pm 0.012 & 0.272\pm 0.029 \\
  \end{array}
\end{equation}

\begin{equation}\label{eq:th15}
  \rowcolors{2}{gray!45!white}{gray!20!white}
  \begin{array}{c|c|c|c}
    a & \rm{ANTONYM} & \rm{NONE} & \rm{SYNONYM} \\
    \hline
    1 & 0.058\pm 0.039 & 0.178\pm 0.011 & 0.259\pm 0.027 \\
    0.8 & 0.086\pm 0.039 & 0.168\pm 0.010 & 0.273\pm 0.026 \\
    0.5 & 0.081\pm 0.037 & 0.169\pm 0.010 & 0.249\pm 0.026 \\
    0.2 & 0.055\pm 0.038 & 0.193\pm 0.010 & 0.241\pm 0.027 \\
    0 & 0.121\pm 0.044 & 0.214\pm 0.011 & 0.259\pm 0.027 \\
  \end{array}
\end{equation}

As can be seen from the previous tables, the pattern \eqref{ANS} is still respected, just as one would expect considering the good Gaussianity results found in Section \ref{sec:gaussianity}.

\subsection{Matrix and vector baselines}\label{sec:baselines} 

\subsubsection{Matrix baseline}

In this section we perform a direct comparison of  the matrices which provide our representation of verbs instead of converting them into observable-value or observable-deviation vectors. This comparison is done through the normalised scalar product
\begin{equation}
  M_1 \cdot M_2 = \frac{\rm{Tr} \left[ M_1 \, M_2^T\right]}{\sqrt { \rm{Tr} \left[ M_1 \, M_1^T\right] \, \rm{Tr} \left[M_2 \, M_2^T\right] } } =\frac{V_1 \cdot V_2}{\sqrt { V_1^2 \, V_2^2} } \> ,
\end{equation}
were we defined the vectors $V^A$ as
\begin{equation}
  V^A = \begin{pmatrix}
    M^A_{11}, & \ldots, & M^A_{1n}, & M^A_{21}, & \ldots, & M^A_{2n}, & \ldots, & M^A_{n1}, & \ldots, & M^A_{nn} 
  \end{pmatrix}^T\> .
\end{equation}
Considering the usual five values for the parameter $a$ in \eqref{eq:mixing}, we find
\begin{equation}
  \rowcolors{2}{gray!45!white}{gray!20!white}
  \begin{array}{c|c|c|c}
    a & \rm{ANTONYM} & \rm{NONE} & \rm{SYNONYM} \\
    \hline
    1 & 0.216\pm 0.009 & 0.134\pm 0.002 & 0.196\pm 0.006 \\
    0.8 & 0.197\pm 0.012 & 0.100\pm 0.002 & 0.176\pm 0.007 \\
    0.5 & 0.205\pm 0.011 & 0.114\pm 0.002 & 0.184\pm 0.007 \\
    0.2 & 0.215\pm 0.009 & 0.131\pm 0.002 & 0.194\pm 0.006 \\
    0 & 0.196\pm 0.012 & 0.099\pm 0.002 & 0.175\pm 0.007 \\
  \end{array}
\end{equation}
As a further baseline check we directly compare the cosine distance of the deviation vectors $V_A$, in other words we compute the cosine distance of deviation vectors
as explained in Section \ref{sec:defs} but using $v^A = V^A$ instead of $v^A_{\alpha}  = \cO_{\alpha }  ( M^A )$. This analysis gives
\begin{equation}
  \rowcolors{2}{gray!45!white}{gray!20!white}
  \begin{array}{c|c|c|c}
    a & \rm{ANTONYM} & \rm{NONE} & \rm{SYNONYM} \\
    \hline
    1 & 0.095\pm 0.010 & 0.047\pm 0.002 & 0.114\pm 0.006 \\
    0.8 & 0.113\pm 0.012 & 0.053\pm 0.002 & 0.129\pm 0.007 \\
    0.5 & 0.111\pm 0.011 & 0.053\pm 0.002 & 0.128\pm 0.007 \\
    0.2 & 0.099\pm 0.010 & 0.048\pm 0.002 & 0.118\pm 0.006 \\
    0 & 0.113\pm 0.012 & 0.053\pm 0.002 & 0.129\pm 0.007 \\
  \end{array}
\end{equation}

As can be seen from the above results, none of these more direct methods provides us with the same sort of hierarchical relation \eqref{ANS}, which further strengthens the intuition that observable-deviation vectors actually capture relevant semantic information.

\subsubsection{Vector baseline from word2vec}

In a very similar vein, one can perform synonym/antonym comparison with word2vec vectors, by directly building permutation invariant functions for these. The observables in this case are 
\bea 
&& \sum_i v_i  \cr 
&& \sum_{ i } v_i^2  , \sum_{ i , j } v_i v_j \cr 
&& \sum_{ i } v_i^3 , \sum_{ i_1 , i_2 } v_{ i_1}^2 v_{i_2} , \sum_{ i_1 , i_2 , i_3 } v_{ i_1 } v_{ i_2} v_{ i_3 } \cr 
&& \sum_{ i } v_i^4 , \sum_{ i_1 , i_2 } v_{ i_1}^3 v_{ i_2} , \sum_{ i_1, i_2 } v_{i_1}^2 v_{ i_2}^2 , 
\sum_{ i_1 , i_2 , i_3 } v_{ i_1}^2 v_{ i_2} v_{ i_3} , \sum_{ i_1 , i_2 , i_3 , i_4} v_{ i_1 } v_{ i_2} v_{ i_3} v_{ i_4}  \> ,  
\eea
where a similar enumeration can be easily done for even higher orders: in general, for degree $n$ the number of observables is $p(n)$, the number of partitions of $n$ -- i.e. the number of ways of writing $n$ as a sum of positive integers. 

Using the above observables with set1 being linear+quadratic, set2 being cubic + quartic and set3 being the full set of observables, the cosine distances one gets are

\begin{equation}
  \rowcolors{2}{gray!45!white}{gray!20!white}
  \begin{array}{c|c|c|c}
     & \rm{ANTONYM} & \rm{NONE} & \rm{SYNONYM} \\
    \hline
    \rm{set1} &  0.397 \pm 0.044 &  0.180 \pm   0.012 & 0.372 \pm  0.028  \\
    \rm{set2} & 0.526 \pm 0.046 & 0.270 \pm 0.012 & 0.460 \pm  0.029  \\
    \rm{set3} & 0.461 \pm 0.043 &  0.223 \pm   0.012 & 0.414 \pm  0.027  \\
  \end{array}
\end{equation}

Next we take directly the input vectors of word2vec, which we call again $V^A$, and set $v^A=V^A$. Using the plain
scalar product (normalised by the module of the individual vectors) and then instead the cosine distance of the deviation vectors we have
\begin{equation}
  \rowcolors{2}{gray!45!white}{gray!20!white}
  \begin{array}{c|c|c|c}
     & \rm{ANTONYM} & \rm{NONE} & \rm{SYNONYM} \\
    \hline
    \rm{plain} &  0.364 \pm 0.015 &  0.268 \pm   0.003 & 0.391 \pm  0.010  \\
    \rm{cosine} & 0.235 \pm 0.017 & 0.137 \pm 0.003 & 0.288 \pm  0.011  \\
  \end{array}
\end{equation}

Neither of the above reproduces the nice ordering of \eqref{ANS}. The failure of the plain inner product is expected since antonyms and synonyms often appear in the same contexts and word2vec learns the word vectors based on co-occurrence in context. For the cosine of observable deviation vectors constructed permutation invariant functions of the vectors, the failure suggests that the success of the observable deviation vectors for matrices relies both on the notion of vectors of permutation invariants and on the representation of words by matrices. We aim to investigate this further in the future.

\section{ Concepts and geometries: Balanced accuracies for  lexical relation tasks }\label{sec:precision}

Having shown in section \ref{sec:relations} that in our construction of observable deviation vectors, once we separate the dataset basing on lexical relations, the means  satisfy the specific ordering \eqref{ANS} we now turn to the task of using this information for classification purposes. More specifically,
we use the values of the means and standard deviations of the cosine distance computed from the experimental data in the previous section to fix a divide between the different word categories. For example, considering synonyms and antonyms we call their means $D_S$ and $D_A$ respectively, and their standard deviations $\Delta_S$ and $\Delta_A$. We then define the divide $D_{A/S}$ as 
\begin{equation}\label{eq:divideAS}
  D_{A/S} = D_A + \frac{4}{\pi}\arctan{\left(\frac{ \Delta_A }{ \Delta_S }\right)} \frac{( D_S - D_A )}{2} \>,
\end{equation}
which can then be used to classify word pairs from their cosine distance by declaring those falling to the right of $D_{A/S}$ as synonyms and those falling to the left as antonyms, as schematically represented in Figure \ref{fig:AS}.
The function $ \frac{4}{\pi}\arctan(x)$  takes the value $1$ when $x=1$, the value $0$ at $x=0$ and tends to 
$2$ as $x$ tends to $\infty$.  This means that when the standard deviations are equal, the divide is half-way between the means.  When the ratio $\Delta_A/\Delta_S$  is very small then the divide is very close to $D_A$ and when  the ratio is very large the divide is very close to $D_S$. In practice, for the experiments in this paper, the ratios are reasonably close to $1$ and we can equally well choose just the ratio $ \Delta_A/\Delta_S$ in place of 
$\frac{4}{\pi}\arctan{\left(\frac{ \Delta_A }{ \Delta_S }\right)}$ to get comparable results. The idea of having a sensible divide which separates the range of cosines into disjoint intervals for different concepts - here synonymy and antonymy - can be viewed as a simple application of the idea of convex regions corresponding to concepts in low dimensional spaces \cite{Garden}. 

\begin{figure}[h]
  \centering
  \begin{tikzpicture}[scale=15]
  
    \draw (-8pt,0) -- (8pt,0);
    \draw (-4pt,0.4pt) -- (-4pt,-0.4pt);
    \draw (4pt,0.4pt) -- (4pt,-0.4pt);
    \draw (0,-0.7pt) -- (0,0.7pt);
    \draw[dashed] (-8pt,0) -- (-10pt,0);
    \draw[dashed] (8pt,0) -- (10pt,0);

    \node at (-4pt,-1pt) {$D_{\rm A}$};
    \node at (4pt,-1pt) {$D_{\rm S}$};
    \node at (0,2.5pt) {$\underbrace{D_A + \frac{4}{\pi}\arctan{\left(\frac{ \Delta_A }{ \Delta_S }\right)} \frac{( D_S - D_A )}{2}}$};

    \draw[->,red,>=Stealth] (-0.2pt,-2pt) -- (-8pt,-2pt);
    \node at (-4pt,-3pt) {\textcolor{red}{predicted antonyms}};

    \draw[->,blue,>=Stealth] (0.2pt,-2pt) -- (8pt,-2pt);
    \node at (4pt,-3pt) {\textcolor{blue}{predicted synonyms}};

  \end{tikzpicture}
  \caption{Figure illustrating antonym-synonym separation criterion. }
\label{fig:AS}
\end{figure}

\noindent
As a measure of how well our criterion is performing we chose to use the balanced accuracy. In order to relate to standard terminology (see for example \cite{Wikiprecrec})  we consider the binary classification problem with the two classes being given by positive (synonyms) and negative (antonyms), then following the notation of Table \ref{tab:confusionmat} we can define the True Positive Rate (TPR) and True Negative Rate (TNR) as
\begin{equation}
  \text{TPR}=\frac{\rm TP}{ \rm TP+FN} \>, \hspace{1cm} \text{TNR}=\frac{ \rm TN}{\rm TN+FP} \>.
\end{equation}
Translating back into the terminology of synonyms and antonyms 
\bea 
&&  \text{TPR} = { { \rm Correctly ~  predicted ~ synonyms} \over { \rm Total ~  number~  of ~  synonyms ~ in ~ dataset } } \cr 
 &&  \text{TNR} =  { { \rm Correctly ~  predicted ~ antonyms } \over { \rm Total ~ number ~ of ~ antonyms ~ in~ dataset } }
\eea
In a machine learning context the TPR is also called Recall and gives an estimate of how many of the total positive cases have been correctly identified. As an intuitive measure of performance over both the positive and negative classes one can then define the accuracy
\begin{equation}
  \text{Accuracy} = \frac{\rm TP + TN}{\rm TP + FN + TN + FP} \>,
\end{equation} 
which has however the drawback of being imbalanced, in the sense that it gives misleading estimates when the sample size of positive terms is different from that of negative terms\footnote{For example, if the ratio of synonyms:antonyms is 3:1, as is roughly the case for the SimVerb-3500 dataset, then even if we classified incorrectly all the antonyms we would still get 75\% accuracy.}. So instead we use Balanced Accuracy which is given by 
\begin{equation}\label{eq:balancedacc}
  \text{Balanced Accuracy}= \frac{\rm TPR + TNR}{2} \>.
\end{equation}

\begin{table}
  \begin{center}
\begin{tabular}{c|c|c|c}
 \multicolumn{2}{c|}{} & \multicolumn{2}{c}{Predicted condition} \\
 \cline{3-4}
 \multicolumn{2}{c|}{} & Positive & Negative \\
\hline
\multirow{2}{4em}{Actual condition} & Positive & True Positive (TP) & False Negative (FN) \\
\cline{2-4}
& Negative & False Positive (FP) & True Negative (TN) 
\end{tabular}
\caption{Contingency table of the outcomes in a generic binary classification problem. In our case we can think of positive=synonym and negative=antonym.}
\label{tab:confusionmat}
\end{center}
\end{table}

\noindent
In general we will consider two different cases:
\begin{itemize}
  \item \textbf{Full set} The full set of data at our disposal is used to determine the location of the divide $D_{A/S}$, in other words $D_{A/S}$ is determined from the means and standard deviations computed over all the synonym and antonym pairs available from SimVerb-3500. Furthermore, we then take the same data set and use the criterion above to distinguish between synonyms and antonyms (positive and negative) and compute the Balanced Accuracy.
  \item \textbf{Partial set} We use a subset corresponding to 65\% of the total synonym/antonym pairs to determine the divide $D_{A/S}$, and then take the complementary data set and try to predict the lexical relation between pairs. The Balanced Accuracy is thus compute on this complementary data set of size 35\% of the total set. In order to avoid biases due to the specific choice of the subsets this procedure is iterated 20 times randomly splitting the pairs into ``training'' and ``test'' set, and the final result is obtaining by averaging. This iterative procedure also allows us to associate an error bar to the averages. This is computed as the standard deviation and displayed alongside the averages in the tables.
\end{itemize}

In this section we consider observable-deviation vectors for the 13, 15 and 28 observable cases and use the normalised product as defined in \eqref{ObsDevSig}.

\subsection{Balanced accuracy for Antonym/Synonym distinction}

Given the definitions above we can compute the Balanced Accuracy for the synonym/antonym distinction, in the two cases described in the previous section we find

\begin{itemize}
  \item {\bf Full set:}

  \begin{equation}\label{eq:recallsynant}
    \rowcolors{2}{gray!45!white}{gray!20!white}
    \begin{array}{c|c|c|c|c|c}
      & 1 & 0.8 & 0.5 & 0.2 & 0 \\
      \hline
      \text{13 obs} & 0.56 & 0.557 & 0.575 & 0.53 & 0.523 \\
      \text{15 obs} & 0.579 & 0.574 & 0.564 & 0.571 & 0.554 \\
      \text{28 obs} & 0.555 & 0.578 & 0.587 & 0.575 & 0.564 \\  \end{array}
  \end{equation}

  \item {\bf Partial set:}
  \begin{equation}\label{eq:recsynantpartial}
    \rowcolors{2}{gray!45!white}{gray!20!white}
    \resizebox{0.85\hsize}{!}{$
    \begin{array}{c|c|c|c|c|c}
      & 1 & 0.8 & 0.5 & 0.2 & 0 \\
      \hline
      \text{13 obs} & 0.569 \pm 0.005 & 0.564 \pm 0.009 & 0.576 \pm 0.008 & 0.548 \pm 0.009 & 0.543 \pm 0.008 \\
      \text{15 obs} & 0.577 \pm 0.009 & 0.573 \pm 0.007 & 0.58 \pm 0.009 & 0.581 \pm 0.009 & 0.559 \pm 0.008 \\
      \text{28 obs} & 0.561 \pm 0.007 & 0.579 \pm 0.008 & 0.597 \pm 0.007 & 0.591 \pm 0.008 & 0.577 \pm 0.010 \\
 
    \end{array}$}
  \end{equation}

\end{itemize}

In the second case, where the divide is defined on a subset of the dataset and the balanced accuracy is computed on the complementary set, we remind the reader that the error comes from averaging over 20 different ``training'' and ``test'' sets.   Alongside the means, we have quoted the standard errors of these means \cite{Wiki-standard-error}  which is calculated by dividing the standard deviation obtained from the variation of the means across the training sets with the square root of 20. The differences between these success rates and the random success rate of 0.500 are large multiples of the standard error, so clearly statistically significant. For example  the lowest success rate of 0.543  (for 13 observables with $a=0$) differs from the random success rate of 0.500 by 0.043 which is 5.4 times the standard error. The highest success rate of 0.597 (28 observables with $a=0.5$) differs from 0.500 by 0.097 which is 13.6 times the standard error of 0.007. It is worth stressing that this is a proof of concept, we see that permutation invariant observables carry relevant information in a linguistic context but our method is far from optimised for NLP tasks.
An example of a possible refinement would be to assign arbitrary weights to the different terms corresponding to different observables in the sum over $\alpha $ in \eqref{ObsDevSig}.  The weights would to be learned from data by requiring the cosine distance of synonym pairs to tend to 1 and those of antonym pairs to tend to -1. 
We leave future refinements along these lines for interesting future investigations.

\subsection{Balanced accuracy including  no-relation pairs } \label{sec:distributions}

An extension of the classification task presented in the previous section is given by the inclusion of word pairs with no lexical relation, to which we refer as None pairs (using the labelling in SimVerb3500). The generalisation of the definition of balanced accuracy beyond the binary classification case is simple and the geometric interpretation of the divide easily extended. We give a schematic representation of the geometry in figure \ref{fig:ANS}, where $D_S$, $D_N$, $D_A$ and $\Delta_S$, $\Delta_N$, $\Delta_A$ are respectively the mean and standard deviation of the Synonym, None and Antonym pairs, and $D_{A/N}$ and $D_{N/S}$ are defined as
\begin{equation}
  D_{A/N} = D_{\rm A} + \frac{4}{\pi}\arctan{\left(\dfrac{\Delta_{\rm A}}{\Delta_{\rm N}}\right)}\dfrac{(D_{\rm N}-D_{\rm A})}{2}
\end{equation}
and
\begin{equation}
  D_{N/S} = D_{\rm N} + \frac{4}{\pi} \arctan{\left(\dfrac{\Delta_{\rm N}}{\Delta_{\rm S}}\right)}\dfrac{(D_{\rm S}-D_{\rm N})}{2} \> .
\end{equation}
To definition of  balanced accuracy uses 
\bea 
&& { \rm True ~ synonym~  rate } =  { { \rm Correctly ~  predicted ~ synonyms} \over { \rm Total ~  number~  of ~  synonyms ~ in ~ dataset } } \cr 
 &&  \text{True ~no ~ relation rate } =  { { \rm Correctly ~  predicted ~ no~ relation~ pairs  } \over { \rm Total ~ number ~ of ~ no ~relation ~ pairs ~ in~ dataset } }  \cr 
 && { \rm True ~ antonym~   rate } =  { { \rm Correctly ~  predicted ~ antonyms } \over { \rm Total ~  number~  of ~  antonyms ~ in ~ dataset } } \cr 
 && 
\eea
The balanced accuracy is then 
\bea 
{ \rm Balanced ~ accuracy } = { 1 \over 3 } \left ( { \rm True ~  synonym ~ rate }   +  \text{True ~ no-relation ~ rate }  
+ { \rm True ~  antonym ~  rate } \right ) \cr 
&& 
\eea

\begin{figure}[h]
  \centering
  \begin{tikzpicture}[scale=15]
  
    \draw (-11pt,0) -- (11pt,0);
    \draw (-4.5pt,0.7pt) -- (-4.5pt,-0.7pt);
    \draw (4.5pt,0.7pt) -- (4.5pt,-0.7pt);
    \draw (0,-0.4pt) -- (0,0.4pt);
    \draw (-7.5pt,-0.4pt) -- (-7.5pt,0.4pt);
    \draw (7.5pt,-0.4pt) -- (7.5pt,0.4pt);
    \draw[dashed] (-11pt,0) -- (-13pt,0);
    \draw[dashed] (11pt,0) -- (13pt,0);

    \node at (-7.5pt,-1pt) {$D_{\rm A}$};
    \node at (0pt,-1pt) {$D_{\rm N}$};
    \node at (7.5pt,-1pt) {$D_{\rm S}$};
    \node at (-4.5pt,1.7pt) {$D_{A/N}$};
    \node at (4.5pt,1.7pt) {$D_{N/S}$};

    \draw[->,red,>=Stealth] (-4.7pt,-2pt) -- (-11pt,-2pt);
    \node at (-8pt,-3pt) {\textcolor{red}{predicted antonyms}};

    \draw[->,blue,>=Stealth] (4.7pt,-2pt) -- (11pt,-2pt);
    \node at (8pt,-3pt) {\textcolor{blue}{predicted synonyms}};

    \draw[<->,green!55!black,>=Stealth] (-4.3pt,-2pt) -- (4.3pt,-2pt);
    \node at (0pt,-3pt) {\textcolor{green!55!black}{predicted none}};

  \end{tikzpicture}
  \caption{Figure illustrating antonym-synonym separation criterion }%
\label{fig:ANS}
\end{figure}

For concreteness, considering the full dataset to compute the divides as well as the Balanced Accuracy, one finds
\begin{equation}
  \rowcolors{2}{gray!45!white}{gray!20!white}
  \begin{array}{c|c|c|c|c|c}
    & 1 & 0.8 & 0.5 & 0.2 & 0 \\
    \hline
    \text{13 obs} & 0.36 & 0.376 & 0.379 & 0.356 & 0.346 \\
    \text{15 obs} & 0.382 & 0.374 & 0.366 & 0.384 & 0.37 \\
    \text{28 obs} & 0.367 & 0.387 & 0.384 & 0.372 & 0.374 \\
  \end{array}
\end{equation}
or alternatively, averaging over different training/test sets, we have
\begin{equation}
  \rowcolors{2}{gray!45!white}{gray!20!white}
  \resizebox{0.85\hsize}{!}{$
  \begin{array}{c|c|c|c|c|c}
    & 1 & 0.8 & 0.5 & 0.2 & 0 \\
    \hline
    \text{13 obs} & 0.369 \pm 0.004 & 0.378 \pm 0.004 & 0.385 \pm 0.005 & 0.364 \pm 0.005 & 0.359 \pm 0.005 \\
    \text{15 obs} & 0.384 \pm 0.006 & 0.375 \pm 0.005 & 0.378 \pm 0.006 & 0.39 \pm 0.006 & 0.374 \pm 0.005 \\
    \text{28 obs} & 0.372 \pm 0.005 & 0.387 \pm 0.005 & 0.391 \pm 0.006 & 0.39 \pm 0.006 & 0.379 \pm 0.006 \\

  \end{array}$}
\end{equation} 
These figures are better than random guessing which would give a success rate of  0.333 for random guesses of one out of three categories (synonym/no-relation/antonym) for each pair of verbs.  The highest success rate  of 0.391 occurs for 28 observables using $a =0.5$. In this case the difference from the random-guessing success rate is 
$0.391-0.333 = 0.058$ which is $9.7$ times the standard error. The minimum success rate of $0.359$ occurs for 13 observables with $a=1$, in which case the difference of $0.359-0.333 = 0.026$ is 5.2 times the standard error. 
While the success rates in this task are significantly different in a statistical sense from the random guessing rate, they are probably not high enough to be competitive  for practical applications in Artificial Intelligence tasks. The relatively low success rates in this sense is explained by the fact that the means of the three distributions are relatively close with standard deviations of the same order, which translates into the fact the window between the divide $D_{A/N}$ and $D_{N/S}$ characterising the no-relation is rather small. This in turn leads to correct identification of the no-relation pairs only around $\sim$5\% of the times, and this reduces the Balanced Accuracy drastically compared to the scenario where only synonyms and antonyms had been considered.
On the other hand, looking at the normalised distributions of the word pairs (Figure \ref{fig:distributions1}) one could have expected that the cosine distance alone is in general not enough for the correct identification of the lexical relations when also the None pairs are included. The cosine distances for None pairs are distributed in approximately uniform way, as opposed for example to the Synonym pairs whose numbers clearly increase as we approach higher values of the normalised cosine distance. So while using \eqref{eq:divideAS} to distinguish between Antonyms and Synonyms appears reasonable based on the data plotted in Figure \ref{fig:distributions1}, it is also clear how further input would be needed for a consistent distinction between for example Antonyms and None.

Alternatively, one could try to single out Synonyms among word pairs of several lexical types like for example a set of Synonyms, Antonyms and None pairs. Here we use again \eqref{eq:divideAS} to set the divide and \eqref{eq:balancedacc} to measure performance, and simply identify positive = Synonyms and negative = (Antonyms $\cup$ None). This can be expected to work reasonably well. Indeed the results are not far from the case where we are only considering the separation of  Antonyms and Synonyms: see \eqref{eq:synvsnonsyn} and compare to \eqref{eq:recallsynant}. It is evident that  trying to single out the Antonyms instead would be less effective due to the form of the distribution  of the None pairs seen in the histograms (Figure \ref{fig:distributions1}). 

\begin{equation}\label{eq:synvsnonsyn}
  \rowcolors{2}{gray!45!white}{gray!20!white}
  \begin{array}{c|c|c|c|c|c}
    & 1 & 0.8 & 0.5 & 0.2 & 0 \\
    \hline
    \text{13 obs} & 0.551 & 0.568 & 0.571 & 0.555 & 0.55 \\
    \text{15 obs} & 0.522 & 0.528 & 0.524 & 0.524 & 0.508 \\
    \text{28 obs} & 0.54 & 0.556 & 0.554 & 0.557 & 0.554 \\ 
  \end{array}
\end{equation}

\section{Hyper/hyponym vs cohyponyms}\label{sec:hyper}

Beyond the lexical classes of Synonyms, Antonyms and unrelated (no-relation) word pairs, SimVerb3500 also contains a set of 190 CoHyponym pairs and 800 Hyper/Hyponym pairs. In this section we show that the cosine distance of observable deviation vectors also  provides information also about these word classes.

\subsection{Mean cosines  of observable deviation vectors}

First we consider the cosine distance of observable deviation vectors as computed in \eqref{eq:cosdis} for the Hyper/Hyponym and CoHyponym pairs. As done in Section \ref{sec:relations}, we consider multiple relevant sets of observables, in particular 13, 28 and 15 and for each we consider four different matrix representations corresponding to different values of the parameter $a$ in \eqref{eq:mixing}. Alongside the mean values, we quote the standard errors of these means \cite{Wiki-standard-error}. In the present case, these are standard deviations divided by the the square root of the number of hyper/hyponym pairs (800) and CoHyponym pairs (190) in the sample respectively.
For 13 variables we have:
\begin{equation}
  \rowcolors{2}{gray!45!white}{gray!20!white}
  \begin{array}{c|c|c}
    & \rm{HYPER/HYPONYM} & \rm{COHYPONYMS}  \\
    \hline
    1 & 0.128 \pm 0.018 & 0.183 \pm 0.035 \\
0.8 & 0.134 \pm 0.018 & 0.194 \pm 0.034 \\
0.5 & 0.137 \pm 0.018 & 0.219 \pm 0.036 \\
0.2 & 0.125 \pm 0.018 & 0.24 \pm 0.037 \\
0 & 0.125 \pm 0.018 & 0.24 \pm 0.037 \\
  \end{array}
\end{equation}
\noindent
For 15 variables: 

\begin{equation}
  \rowcolors{2}{gray!45!white}{gray!20!white}
  \begin{array}{c|c|c}
    & \rm{HYPER/HYPONYM} & \rm{COHYPONYMS}  \\
    \hline
    1 & 0.234 \pm 0.017 & 0.315 \pm 0.037 \\
0.8 & 0.233 \pm 0.017 & 0.333 \pm 0.035 \\
0.5 & 0.239 \pm 0.019 & 0.339 \pm 0.039 \\
0.2 & 0.298 \pm 0.02 & 0.416 \pm 0.043 \\
0 & 0.323 \pm 0.02 & 0.447 \pm 0.041 \\
  \end{array}
\end{equation}
\noindent
For 28 observables:
\begin{equation}
  \rowcolors{2}{gray!45!white}{gray!20!white}
  \begin{array}{c|c|c}
    & \rm{HYPER/HYPONYM} & \rm{COHYPONYMS}  \\
    \hline
    1 & 0.133 \pm 0.016 & 0.204 \pm 0.033 \\
0.8 & 0.141 \pm 0.016 & 0.209 \pm 0.033 \\
0.5 & 0.145 \pm 0.017 & 0.231 \pm 0.034 \\
0.2 & 0.139 \pm 0.017 & 0.254 \pm 0.035 \\
0 & 0.139 \pm 0.017 & 0.256 \pm 0.035 \\
  \end{array}
\end{equation}

It is interesting to notice that  a hierarchical relation similar to \eqref{ANS} also holds for the means of the Hyper/Hyponym and CoHyponym pairs:
\begin{equation}
  \text{HYPER/HYPONYM} < \text{COHYPONYMS} \> .
\end{equation}
The separation of the means are typically several times the standard errors in these means. For example, for 28 observables in the last table, choosing $a=1$, we see a separation $0.204 -0.133 = 0.71$ which is 21.5 times the larger of the standard errors ($0.033$). The robustness of the hierarchy, with respect to changes of the $a$-parameter or changes of the set of observables, motivates the formulation of a geometrical approach to the distinction of hyper/hyponym pairs from co-hyponyms. The success rates of the approach are sensitive to the more detailed properties of the distributions such as the standard deviations. The geometrical approach to this distinction task and the success rates are detailed in Section \ref{BAHYCOH}.  

  While both the CoHyponyms as well as Hyper/Hyponym pairs can be thought of presenting some sort of similarity, and thus one could expect a certain degree of alignment in the associated vectors in analogy to the Synonym case, the CoHyponym pairs seem to consistently present larger average  values for the cosines. In a hypernym/hyponym pair such as (animal, dog) the hypernym describes a wider class. In a co-hyponym pair such as (cat,dog) the two words are in the same class.   It may seem plausible that co-hyponyms are closer to each other in an intuitive sense than hypernym/hyponym pairs.  Nevertheless, this intuition contradicts with the human similarity judgements collected for SimVerb 3500, where an average of 4.43 for co-hyponyms and 6.01 for hyper/hyponyms is reported.  Recent work has shown that word vector spaces can be transformed into spaces whose geometry reflects the hyper/hypo relation  \cite{VuMr2018}. We think a similar transformation can be learnt for co-hyponym pairs. It would also be interesting to think about ways of reformulating the semantic similarity question such that human annotators produce judgements that correlate with it.

\subsection{Balanced accuracy on hyper/hyponym vs cohyponym pair distinction}\label{BAHYCOH}

We perform the same two balanced accuracy computations as discussed in Section \ref{sec:precision},
but this time with the Hyper/Hyponym pairs taking on the role of the Antonyms and the CoHyponym pairs taking the role of the Synonyms. A graphical representation is given in Figure \ref{fig:HHC}, where $D_H/\Delta_H$ and $D_C/\Delta_C$ are the mean/standard deviation of the Hyper/Hyponym and CoHyponym pair respectively.

\begin{figure}[h]
  \centering
  \begin{tikzpicture}[scale=15]
  
    \draw (-8pt,0) -- (8pt,0);
    \draw (-4pt,0.4pt) -- (-4pt,-0.4pt);
    \draw (4pt,0.4pt) -- (4pt,-0.4pt);
    \draw (0,-0.7pt) -- (0,0.7pt);
    \draw[dashed] (-8pt,0) -- (-10pt,0);
    \draw[dashed] (8pt,0) -- (10pt,0);

    \node at (-4pt,-1pt) {$D_{\rm H}$};
    \node at (4pt,-1pt) {$D_{\rm C}$};
    \node at (0,2.5pt) {$\underbrace{D_{\rm H} + \frac{4}{\pi}\arctan{\left(\dfrac{\Delta_{\rm H}}{\Delta_{\rm C}}\right)}\dfrac{(D_{\rm C}-D_{\rm H})}{2}}$};

    \draw[->,red,>=Stealth] (-0.2pt,-2pt) -- (-8pt,-2pt);
    \node at (-4pt,-3pt) {\textcolor{red}{Hyper/Hyponym}};

    \draw[->,blue,>=Stealth] (0.2pt,-2pt) -- (8pt,-2pt);
    \node at (4pt,-3pt) {\textcolor{blue}{CoHyponyms}};

  \end{tikzpicture}
  \caption{Figure illustrating antonym-synonym separation criterion }%
\label{fig:HHC}
\end{figure}

We find the followings values: 
\begin{itemize}
  \item {\bf Full set:}
  

  \begin{equation}
    \rowcolors{2}{gray!45!white}{gray!20!white}
    \begin{array}{c|c|c|c|c|c}
      & 1 & 0.8 & 0.5 & 0.2 & 0 \\
      \hline
      \text{13 obs} & 0.525 & 0.529 & 0.537 & 0.556 & 0.551 \\
      \text{15 obs} & 0.533 & 0.524 & 0.536 & 0.547 & 0.545 \\
      \text{28 obs} & 0.531 & 0.543 & 0.528 & 0.56 & 0.566 \\ 
    \end{array}
  \end{equation}

  \item {\bf Partial set:}

\begin{equation}\label{HHpart}
  \rowcolors{2}{gray!45!white}{gray!20!white}
  \resizebox{0.85\hsize}{!}{$
    \begin{array}{c|c|c|c|c|c}
      & 1 & 0.8 & 0.5 & 0.2 & 0 \\
      \hline
      \text{13 obs} & 0.517 \pm 0.005 & 0.52 \pm 0.005 & 0.535 \pm 0.005 & 0.547 \pm 0.004 & 0.545 \pm 0.006 \\
      \text{15 obs} & 0.527 \pm 0.004 & 0.524 \pm 0.005 & 0.529 \pm 0.006 & 0.538 \pm 0.006 & 0.539 \pm 0.006 \\
      \text{28 obs} & 0.527 \pm 0.005 & 0.536 \pm 0.005 & 0.528 \pm 0.006 & 0.56 \pm 0.005 & 0.563 \pm 0.006 \\
 
    \end{array}$}
\end{equation}

\end{itemize}

  Once again we see that the values of the success rates  are above random success rate of $0.500$. 
The variations following from different choices of training set in the construction of \eqref{HHpart}
lead to the standard errors in the success rates. These  errors are evaluated by taking the standard deviations obtained across the choices of training set, divided by the square root of the number of choices which was 20. We see that these errors of the means are small compared to the difference between the means and the random success rates. The highest success rate (from 28 observables and $a=0$) of 0.563 differs from 0.500 by over 10 times the standard error of 0.006.  The lowest success rate of 0.517 (from 13 observables and $a=1$) differs from the random success rate of 
0.500 by 3.4 times the standard error. The variations produced from different choices of $a$-parameter, or different observable sets are all smaller than the difference from the random guessing success rate. It should be noted however that  success rates are lower than those in the Synonym vs Antonym distinction task (equations \eqref{eq:recallsynant}\eqref{eq:recsynantpartial}).

\subsection{Hyper vs hyponym lengths}\label{sec:longhyper}

In \cite{SantusHyper} it is argued that hypernyms are more ``entropic'' compared to hyponyms. This is motivated by the
wider contexts in which the hypernyms appear compared to the Hyponyms, and this was the foundation for their hypernym/hyponym distinction. We propose the heuristic argument  that matrices representing more entropic words will be denser,  with  more numerous matrix entries significantly different from zero. The permutation invariant  matrix observables are constructed as sums over matrix elements. If we regard the different non-vanishing contributions as steps in a random walk, a more entropic matrix will have effectively  more numerous steps, thus leading to   larger deviations from the mean. We thus expect, based on heuristic arguments, that 
the observable deviation vectors associated to the hypernyms should be longer than those associated to the Hyponyms.
We have  tested this expectation \footnote{In order to distinguish between Hypernym and Hyponym in a given pair we use WordNet: we take the first word, iteratively extract all the hyponyms from the word itself and its hyponyms, and then look for the second word of our pair in this list. If we find it then the second word is a hyponym of the first, else the other way around. Due to a bug in WordNet (the word ``inhibit'' has among its hyponyms ``restrain'', but in turn among the hyponyms of ``restrain'' we find ``inhibit'' and this leads to an infinite loop in our Hyper/Hyponym recognition algorithm) we only use 798 of the total 800 Hyper/Hypo pairs for this computation.} on three sets of observables, the full 28 considered as the
most complete set so far, and then two other sets of respectively low/high node observables.
In particular an observable involving $n$ powers of the matrix $M^A$ was considered as high node
if it had at least $n+1$ distinct nodes. The values given in the table below correspond to
the ratios
\begin{equation}
  \text{ratio} \> = \> \frac{\text{\# of pairs for which hyper $>$ hypo}}{\text{all pairs of hyper/hyponyms}}
\end{equation}

\begin{equation}\label{eq:hyperhypolen}
  \rowcolors{2}{gray!45!white}{gray!20!white}
  \begin{array}{c|c|c}
    \textcolor{white}{hhh}\text{28 obs} \textcolor{white}{hhh} & \textcolor{white}{hhh}\text{high node} \textcolor{white}{hhh} & \textcolor{white}{hhh}\text{low node} \textcolor{white}{hhh} \\
    \hline
      0.627 & 0.602 & 0.629
  \end{array}
\end{equation}

Considering in this case the Mahalanobis measure provides a somewhat significant deviation from the diagonal inner product. In fact, doing the length comparison with \eqref{eq:maha} we find
\begin{equation}
  \rowcolors{2}{gray!45!white}{gray!20!white}
  \begin{array}{c|c|c}
    \textcolor{white}{hhh}\text{28 obs} \textcolor{white}{hhh} & \textcolor{white}{hhh}\text{high node} \textcolor{white}{hhh} & \textcolor{white}{hhh}\text{low node} \textcolor{white}{hhh} \\
    \hline
      0.677 & 0.652 & 0.653
  \end{array}
\end{equation}
which has to be compared with \eqref{eq:hyperhypolen}.

We conclude that combining the framework of compositionality and permutation invariant matrix statistics in this paper with the entropic argument gives success rates of $63\% $ and $68 \%$ when we use the observable deviation vectors with diagonal  or Mahalanobis geometry respectively.

\section{ Summary and outlook }\label{sec:sumout} 

In this paper we have employed polynomial  permutation invariant functions of matrix variables (PIMOs), labelled by directed graphs,   in order to construct low-dimensional vector representations for data taking the form of ensembles of matrices. In the present case we have a collection of matrices $\{ M^A_{ ij} \} $ of size $D =100$, i.e the indices $i, j $ run over $ \{ 1, 2, \cdots D\}$. The index $ A$ runs over the set of $827$ distinct  verbs in the SimVerb3500 dataset. This ensemble of matrices was constructed in \cite{CSC2020}. We worked with vectors of dimensions $13$, $ 28 $ and $15$ ( as well as $10$, $23$ as explained in section {\ref{sec:gaussianity}). These choices were motivated by investigations of Gaussianity of the matrix ensembles. The ensemble averages of the $13$ linear and quadratic functions are used to determine the $13$ parameters of the permutation invariant Gaussian matrix model \cite{PIGMM}. For the extra $15$ polynomial functions, we tested the theoretical predictions of the Gaussian model against the ensemble averages. As in \cite{GTMDS} we found very good evidence for the correlated Gaussianity described by the $13$-parameter model.

 We took the natural next step of considering the deviation vectors of 
PIMOs for each member of the ensemble compared to the ensemble averages. We equipped the vector spaces of deviation vectors with a geometry motivated from random matrix theory, consisting of a metric where the different graphs are orthogonal. We also worked with the Mahalanobis metric from statistics. We used these geometries of observable deviation vectors to address a number of disambiguation tasks associated with synonyms, antonyms, hypernyms and hyponyms.  We found  two ordering relations on the lexical semantic labels that match their human annotations. In particular, the average cosines of pairs of our observable-deviation vectors obey the pattern  ANTONYM $<$ NONE $<$ SYNONYM  on the antonyn/synonym/no-relation labels and HYPER/HYPONYM $<$ COHYPONYMS on the hypernym/hyponym and co-hyponym labels. We use these orderings in three classification tasks for pairs of words  and obtain success rates well above random chance,  taking into account our error estimates. For synonym/antonym distinctions, using angles between pairs of words, we had robust  success rates in the range  $ 55-57 \%$ as a number of discrete parameters in the construction were varied. For the task of distinguishing the hypernym from the hyponym in a hypernym-hyponym pair, we used lengths of deviation vectors to obtain   success rates of approximately $62 \% $ using the diagonal  metric, and success rates near $ 67\%$ using the Mahalanobis metric. We consider these results to be significant properties of the statistical physics of natural language data.

Applying our analysis to the word pairs considered in \cite{AntSyn} would allow us to perform a consistent comparison with existing results. Initial steps towards this direction are encouraging despite the small size of the dataset we could analyse: the overlap between the word pairs in \cite{AntSyn} and the matrix data at our disposal is the limiting factor. It would be interesting to extend the construction of matrices in \cite{CSC2020} to the pairs of \cite{AntSyn} in order to get a more robust testing set. The dataset used in this work, as well as other datasets used in synonym/antonym identification tasks contain nouns, as well as verbs and adjectives.   Compositional distributional semantics only argues for matrix representations for words that have a relational role, i.e. adjectives and verbs. Nouns are treated as \emph{atomic} in that they do not relate any of the parts of speech. There is however  a complementary body of work where all words, regardless of their grammatical role, have a matrix representation, see \cite{Bowman}. Working with these matrices in our model and thus extending the model to noun matrices is a direction for future work.

A further interesting future direction is to explore the parameter spaces of the  metrics (variations on the diagonal  and Mahalanobis geometries we have used)  with machine learning techniques in order to improve the success rates in the tasks we have investigated in this paper. Based on the successful results here, it is reasonable to expect that machine learning will lead to visible improvements in the success rates. Another direction is to consider the comparison of observable deviation vectors  for pairs of verbs using two or three dimensional geometries derived from the $13$, $28$ or $15$ dimensional starting points. In the final step of the synonym/antonym comparison task, we used the angles between pairs of vectors. In the final step of the hypernym/hyponym distinction, we used the lengths of pairs of vectors. It is natural to consider, for instance, synonym pairs, antonym pairs, hypernym/hyponym pairs in a two dimensional space consisting of the angles between the pairs as one direction and the  ratio of the two lengths of the pair, taken together to define a three-dimensional space. We would then look for  convex regions in the  two  dimensional spaces associated with the different types of pairs. The use of regions in the one-dimensional space of cosines in the paper, as well as these generalizations,    are aligned with the idea  of associating concepts to geometries, as a way to explain the efficacy of human learning \cite{Garden}.  The discussions in \cite{Garden}  also potentially offer a route towards an  explanation of why simple low-dimensional geometries (derived from higher dimensional vectors of permutation invariant observables by considering the angle between  and the  lengths of pairs of higher dimensional vectors)  can capture our cognitive classifications of lexical semantic types. The simple low-dimensional derived geometries, such as the one-dimensional space of cosines of angles used  in this paper or the generalizations to two or three dimensions by including lengths, have the important mathematical characteristic of low dimensionality. This characteristic is also a property of the currently observable  physical universe,  and  can be viewed as part of the mathematical ecology of the universe that houses the evolution of the human brain. It may be argued that symmetry-based reductions to low-dimensional spaces which capture aspects of  human cognition (such as our comprehension of lexical semantics)  should be expected, since this cognition is  adapted to the analogous structures in the physical universe.

\vskip.5cm 

\begin{center} 
	{ \bf Acknowledgements} 
\end{center} 
SR is supported by the STFC consolidated grant ST/P000754/1 `` String Theory, Gauge Theory \& Duality” and a Visiting Professorship at the University of the Witwatersrand. MS is supported by a Discovery to Use grant from UCL’s EPSRC IAA 2020-22 Impact Acceleration Account and a senior Royal Academy of Engineering research fellowship. We are pleased to acknowledge useful conversations on the subject of this paper with  George Barnes, Martin Benning, Robert de Mello Koch, Alston Misquitta, Adrian Padellaro, Gopal Ramchurn, Axel Rossberg, Anna Rogers, Greg Slabaugh,  Michael Stephanou,  Lewis Sword and  Gijs Wijnholds. This work was supported by the STFC IAA (PRNZZD9R) Flexible Innovation project at QMUL and by the European Union's Horizon 2020 research and innovation programme under the Marie Sk\l{}odowska-Curie grant agreement No.~764850 {\it ``\href{https://sagex.org}{SAGEX}''}.


\begin{thebibliography}{}
\bibitem{EY}
Edelman and Yang, ``Random Matrix Theory and its innovative applications,''
www-math.mit.edu/~edelman/publications/random\_matrix\_theory\_innovative.pdf
\end{thebibliography}
\end{document}